\documentclass{article}

\usepackage{booktabs}
\usepackage{graphicx}
\usepackage{array}      
\usepackage{colortbl}
\usepackage{xcolor}
\usepackage{subfigure}

\definecolor{lightgreen}{RGB}{200,255,200}
\definecolor{lightred}{RGB}{255,200,200}
\definecolor{green}{RGB}{0,150,0}
\definecolor{red}{RGB}{200,0,0}

    \usepackage[preprint,nonatbib]{neurips_2025}

\usepackage[utf8]{inputenc} 
\usepackage[T1]{fontenc}    
\usepackage{url}            
\usepackage{booktabs}       
\usepackage{amsfonts}       
\usepackage{nicefrac}       
\usepackage{microtype}      
\usepackage{xcolor}         

\usepackage[utf8]{inputenc}
\usepackage[T1]{fontenc}
\usepackage{algorithm}
\usepackage{algorithmic}
\usepackage{graphicx}
\usepackage{booktabs}
\usepackage{multirow}
\usepackage{amsmath}
\usepackage{amsfonts}
\usepackage{enumitem}
\usepackage{url} 
\usepackage{xspace}
\usepackage{xcolor}
\usepackage[hyperfootnotes=false]{hyperref}
\usepackage{epigraph}

\usepackage{amsthm}
\theoremstyle{plain}

\definecolor{langblue}{rgb}{0, 0.4, 0.8}
\definecolor{langred}{rgb}{0.81, 0.09, 0.13}
\definecolor{langgreen}{rgb}{0.0, 0.6, 0.3}
\hypersetup{
    colorlinks=true,
    linkcolor=langblue,
    citecolor=langblue,
}

\makeatletter
\AtBeginDocument{%
  \let\oldref\ref
  \renewcommand{\ref}[1]{%
    \hyperref[{#1}]{\underline{\oldref{#1}}}%
  }%
}
\makeatother

\usepackage{titletoc}
\newcommand\DoToC{%
  \startcontents
  \printcontents{}{1}{\textbf{\large Contents of Appendix}\vskip3pt\hrule\vskip5pt}
  \vskip3pt\hrule\vskip5pt
}

\newcommand{\method}{SuperRL\xspace}

\title{SuperRL: Reinforcement Learning with Supervision to Boost Language Model Reasoning}

\author{
Yihao Liu$^1$\thanks{\ \ Work done during internship at Microsoft.} \hspace{0.5em}
Shuocheng Li$^1$\footnotemark[1] \hspace{0.5em}
Lang Cao$^2$\footnotemark[1] \hspace{0.5em}
Yuhang Xie$^1$\footnotemark[1] \hspace{0.5em}\\
\textbf{
Mengyu Zhou$^3$\thanks{\ \ Corresponding author (mezho@microsoft.com).} \hspace{0.5em}
Haoyu Dong$^3$ \hspace{0.5em}
Xiaojun Ma$^3$ \hspace{0.5em}
Shi Han$^3$ \hspace{0.5em}
Dongmei Zhang$^3$} \\
$^1$Peking University \quad
$^2$University of Illinois Urbana-Champaign \quad
$^3$Microsoft
}

\begin{document}

\maketitle

\begin{abstract}
Large language models (LLMs) are increasingly applied to complex reasoning tasks, where rich high-quality offline data—such as expert-annotated solutions—are often available. However, standard reinforcement learning (RL) struggles in sparse-reward settings and fails to fully leverage offline supervision. We propose SuperRL, a unified training framework that adaptively switches between RL and supervised fine-tuning (SFT) . For any data instance where all sampled trajectories yield zero reward, indicating a lack of gradient signal, SuperRL triggers a fallback to supervised fine-tuning using high-quality offline data. Experiments across diverse reasoning benchmarks demonstrate that SuperRL outperforms standard RL, delivering improved sample efficiency, generalization, and robustness.
\begin{figure}[h]
    \centering
    \includegraphics[width=0.8\textwidth]{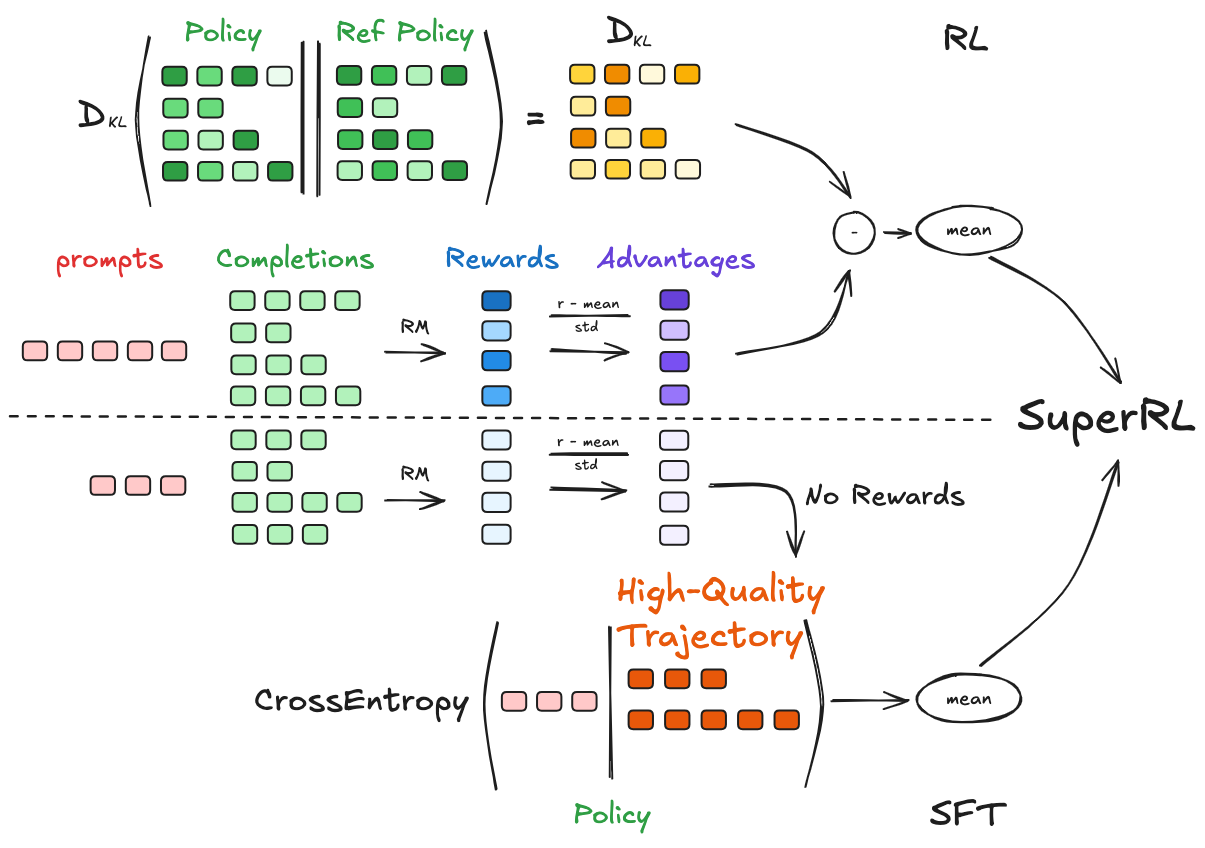}
    \caption{Overview of \textbf{SuperRL}. SuperRL is a unified training framework that adaptively combines RL and SFT based on reward signal. During training, for each input, the model samples multiple rollouts and computes their rewards. If at least one trajectory receives a nonzero reward, standard RL updates are applied using policy gradients. If all trajectories yield zero reward—indicating an absence of learning signal—SuperRL falls back to supervised fine-tuning using high-quality demonstrations.}
    \label{fig:teaser}
\end{figure}
\end{abstract}


\section{Introduction}
LLM reasoning has been studied extensively using lightweight prompting techniques—few-shot examples, chain-of-thought prompting, and self-consistency heuristics—but such static approaches can exhibit diminishing returns in more challenging inference scenarios.~\cite{wei2022chain, wang2022self, brown2020language, wang2022towards, zhang2022automatic, cao2023graphreason} More recently, there has been growing interest in leveraging reinforcement learning for "test-time scaling"~\cite{zhang2025surveytesttimescalinglarge}, empowering models to bolster their reasoning capabilities as they generate outputs. For example, Deepseek R1 demonstrates that applying Group Relative Policy Optimization (GRPO) to fine-tune inference trajectories can lead to more robust and flexible reasoning behaviors, all without modifying the underlying model architecture.~\cite{shao2024deepseekmath, deepseekai2025deepseekr1incentivizingreasoningcapability}

Despite recent progress, online reinforcement learning methods such as PPO and GRPO still face limitations in reasoning tasks.~\cite{shao2024deepseekmath, schulman2017proximalpolicyoptimizationalgorithms} First, these inherently on-policy methods update exclusively from trajectories generated by the current policy and therefore struggle to leverage high-quality offline data—such as human-annotated or distilled reasoning traces—that encode valuable prior knowledge but lie outside the policy distribution.~\cite{levine2020offlinereinforcementlearningtutorial, fujimoto2019offpolicydeepreinforcementlearning} Second, in sparse-reward environments, online rollouts seldom produce enough successful trajectories to supply reliable learning signals, making it difficult to bootstrap coherent reasoning.~\cite{andrychowicz2018hindsightexperiencereplay, ecoffet2021goexplorenewapproachhardexploration} Ultimately, because offline data provide numerous successful trajectories with clear guidance while online rollouts frequently fail and offer little feedback, reinforcement learning struggles to acquire robust reasoning capabilities.

Supervised fine-tuning (SFT) directly leverages offline data with “correct thinking” to teach reliable reasoning paths but lacks any mechanism to learn from mistakes or negative cases, leading it to simply memorize positives ~\cite{wei2022finetunedlanguagemodelszeroshot, sanh2022multitaskpromptedtrainingenables, chu2025sftmemorizesrlgeneralizes}. In contrast, reinforcement learning (RL) can generalize by exploring and learning from failed trajectories ~\cite{chu2025sftmemorizesrlgeneralizes}. While multi-stage SFT + RL—first offline fine-tuning then on-policy RL as in RLHF ~\cite{ziegler2020finetuninglanguagemodelshuman}—anchors models to sound reasoning and adapts them under sparse rewards, it suffers catastrophic forgetting of supervised knowledge during RL and considerable sample and compute inefficiency.~\cite{luo2025empiricalstudycatastrophicforgetting, kotha2024understandingcatastrophicforgettinglanguage, fernando2025mitigatingforgettingllmsupervised} These shortcomings motivate tighter integration—e.g., interleaving or unifying SFT and RL objectives within each update—to preserve alignment, improve stability, and boost efficiency. However, multi-stage SFT+RL pipelines can impair generalization by overfitting to offline traces and narrow RL objectives.~\cite{luo2025empiricalstudycatastrophicforgetting, kotha2024understandingcatastrophicforgettinglanguage, fernando2025mitigatingforgettingllmsupervised}

To enhance language model reasoning across both dense and sparse reward regimes, we propose \textbf{SuperRL}, a unified training framework that dynamically switches between RL and SFT at the instance level. For each training input, the model samples multiple rollouts and evaluates their rewards. If any rollout receives a nonzero reward, policy gradient updates are applied. Otherwise—when all rollouts fail—SuperRL falls back to SFT on expert demonstrations. This simple yet effective fallback mechanism allows the model to benefit from high-quality offline data whenever online learning provides insufficient signal. While online rollouts frequently fail to offer informative supervision, offline datasets contain abundant successful trajectories that are often underutilized. By adaptively leveraging these trajectories, SuperRL closes this gap and facilitates more stable and effective learning. SuperRL is entirely data-driven and achieves strong generalization across reasoning tasks.

\noindent\textbf{Our main contributions are as follows:}
\begin{itemize}
\item \textbf{SFT-guided Reinforcement Learning.} We propose a simple yet powerful mechanism to integrate SFT signals into the RL process. This hybrid approach enables the model to fall back on high-quality offline supervision in situations where reward signals are sparse, delayed, or entirely absent, thereby significantly enhancing the learning signal in complex reasoning environments.

\item \textbf{The {SuperRL} Framework.} We introduce \textbf{SuperRL}, a training framework that dynamically switches between RL and SFT at the instance level. The switching is guided by reward feedback, allowing the system to adaptively determine the most appropriate optimization strategy for each input.

\item \textbf{Comprehensive Empirical Validation.} We conduct extensive experiments across multiple reasoning benchmarks and model scales. Results show that \textsc{SuperRL} outperforms standard RL baselines in most settings, yielding consistent gains in sample efficiency, generalization to out-of-distribution problems, and training stability. Notably, these improvements hold across both synthetic tasks and real-world language reasoning benchmarks, highlighting the robustness and practical effectiveness of the proposed framework.
\end{itemize}


\section{Related Work}
\subsection{Reasoning with LLMs}
LLMs exhibit impressive knowledge but often struggle with complex reasoning tasks (e.g. multi-step math or logical problems) without specialized training. Recent research has shown that incorporating explicit reasoning steps can substantially improve performance. For example, training models to produce chain-of-thought solutions (step-by-step reasoning) enables better arithmetic and logic problem solving. ~\cite{xiong2025minimalistapproachllmreasoning} In fact, fine-tuning a pretrained LM on detailed quantitative reasoning demonstrations (as done in Minerva) led to state-of-the-art results on challenging math benchmarks. These findings underscore that vanilla next-token prediction alone is insufficient for high-level reasoning; additional fine-tuning or feedback signals are needed to guide LLMs in reasoning processes. This has driven interest in post-training LLMs specifically for reasoning capabilities, as seen with recent models like DeepSeek-R1, which explicitly target mathematical reasoning skills
~\cite{xiong2025minimalistapproachllmreasoning, deepseekai2025deepseekr1incentivizingreasoningcapability}
. Such models leverage verifiable rewards (e.g. checking a final answer’s correctness) to refine the reasoning ability of LLMs, pointing to the need for training paradigms beyond standard supervised learning.~\cite{deepseekai2025deepseekr1incentivizingreasoningcapability, mroueh2025reinforcementlearningverifiablerewards}

\subsection{Reinforcement Learning for LLMs}
Reinforcement learning from human feedback~\cite{dong2024rlhfworkflowrewardmodeling}—as popularized by InstructGPT—uses on-policy methods like PPO to fine-tune language models against a learned reward signal, but doing so requires costly fresh sampling, a separate value network and careful KL-penalty tuning, and remains sample-inefficient when rewards are asparse.~\cite{schulman2017proximalpolicyoptimizationalgorithms, crowder2024hindsightexperiencereplayaccelerates, xiong2025minimalistapproachllmreasoning} GRPO simplifies PPO by ditching the critic and computing advantages via within-prompt reward normalization across multiple candidates, yet it still depends on large volumes of model-generated outputs and can wander from the pre-trained behavior in low-reward regimes. These limitations—computational overhead, sample inefficiency and instability under sparse feedback—have spurred interest in leaner or hybrid approaches that retain the simplicity of policy gradients while injecting external guidance to stabilize and steer training.\cite{shao2024deepseekmath, xiong2025minimalistapproachllmreasoning}

However, even with these leaner on‐policy variants, sparse‐reward reasoning still yields too few positive signals for stable gradient estimates, leading to stalled or collapsed training~\cite{andrychowicz2018hindsightexperiencereplay, ecoffet2021goexplorenewapproachhardexploration}.

\subsection{Unifying SFT and RL}
Many methods leverage demonstrations to jump-start reasoning. SFT trains a pre-trained LLM on prompt–solution pairs (human-written or model-verified), teaching explicit reasoning patterns and yielding strong baseline performance.\cite{wen2025lightr1curriculumsftdpo, ouyang2022traininglanguagemodelsfollow} However, pure SFT merely imitates its training data: it cannot exceed the quality or coverage of provided solutions, and gaps in demonstration types or optimal answers limit its generalization.\cite{chu2025sftmemorizesrlgeneralizes}

Recent research has demonstrated that combining supervised fine-tuning (SFT) with reinforcement learning (RL) effectively enhances the capabilities of large language models (LLMs). This approach leverages offline high-quality data to guide initial training and employs reinforcement learning to further align model outputs with desired behaviors. Notably, several recent works, including InstructGPT\cite{ouyang2022traininglanguagemodelsfollow}, DeepSeek-R1\cite{deepseekai2025deepseekr1incentivizingreasoningcapability} and Qwen3\cite{yang2025qwen3technicalreport}, have adopted similar strategies to bolster the reasoning abilities of LLMs. \cite{deepseekai2025deepseekr1incentivizingreasoningcapability, yang2025qwen3technicalreport}

Unlike previous approaches that separate RL and SFT into distinct stages, SuperRL treats them as complementary processes governed by the data itself. It selectively applies offline supervision only when online rollouts lack useful reward signals—achieving more stable optimization and better generalization, particularly in tasks where sparse rewards make RL alone unreliable.

\section{Methodology}
\subsection{Reward-Guided Loss Selection}
We introduce a simple yet effective principle: when any rollout trajectory yields a nonzero reward, we apply a standard policy-gradient update to reinforce beneficial behaviors. However, if \emph{all} trajectories return zero reward—signifying complete failure—the model instead performs a supervised update using offline demonstrations. This fallback mechanism ensures continuous learning even in the absence of meaningful reward signals.

Our approach treats the \emph{presence or absence of reward} as a reliable indicator of learning potential. Rather than relying on hand-tuned heuristics or reward-density estimators, the rollout outcomes themselves determine whether reinforcement learning (RL) or supervised fine-tuning (SFT) is appropriate. This design is lightweight, hyperparameter-free, and operates at the per-instance level.

By selectively activating SFT only on challenging examples, the model allocates RL updates to solvable cases while directing supervision toward failure modes. This dynamic allocation improves training stability and accelerates convergence. On easier datasets, SFT is rarely triggered; under sparse-reward regimes, SFT becomes an automatic fail-safe. The result is a unified learning rule that is both simple and robust—consistently outperforming pure RL across diverse reasoning tasks.

\method embodies this philosophy by integrating RL and SFT into a single reward-aware optimization loop that dynamically adjusts the training signal based on rollout feedback.

\subsection{Reward-Gated Optimization in \textsc{SuperRL}}
\label{sec:superrl_loss}

\paragraph{Notation.}
For a task instance $x\!\in\!\mathcal{X}$ we draw $K$ trajectories
$\tilde y_1,\dots,\tilde y_K\!\sim\!\pi_\theta(\cdot\mid x)$ and obtain rewards
$R(x,\tilde y_k)$.  
A demonstration set
$\mathcal{Y}_{\mathrm{SFT}}(x)=\{y^{*(1)},\dots,y^{*(M)}\}$
is available for imitation.
Define the advantage and importance ratio
\[
A_k = R(x,\tilde y_k)-b(x),\qquad
r_k(\theta)=\frac{\pi_\theta(\tilde y_k\!\mid\!x)}
                 {\pi_{\theta_{\mathrm{old}}}(\tilde y_k\!\mid\!x)}.
\]

\paragraph{Two primitive losses.}
\begin{align}
\mathcal{L}_{\mathrm{SFT}}(\theta;x)
  &= -\frac{1}{M}\sum_{m=1}^{M}\log\pi_\theta\!\bigl(y^{*(m)}\mid x\bigr),
\\[4pt]
\mathcal{L}_{\mathrm{PG}}(\theta;x)
  &= -\frac{1}{K}\sum_{k=1}^{K}g\!\bigl(r_k(\theta),A_k\bigr),
\end{align}
where $g(\cdot,\cdot)$ is any monotone policy-gradient surrogate
(e.g.\ PPO, TRPO, GRPO).

\paragraph{Reward-based soft gate.}
We define a soft gate that checks whether \emph{any} rollout receives nonzero reward:
\[
c(x)=\mathbf{1}\!\bigl(\max_k R(x,\tilde y_k)>0\bigr)
\;
\qquad\text{(triggered if any rollout succeeds)}
\]

Based on this gate, we define the unified objective:
\[
\boxed{\;
\mathcal{L}_{SuperRL}(\theta;x)
  =(1-c(x))\,\mathcal{L}_{\mathrm{SFT}}(\theta;x)
  +      c(x)\,\mathcal{L}_{\mathrm{PG}}(\theta;x)
\;}
\]

If every rollout fails ($c(x)=0$), the model performs supervised imitation using demonstrations; otherwise, it reinforces successful behaviors via policy gradients. This hyperparameter-free switch tightly couples exploration and imitation, providing stable and sample-efficient learning that outperforms pure RL across both dense- and sparse-reward reasoning tasks.

\subsection{Design Variants and Unified Hybrid Objective}
\label{sec:design_variants}

The hard switch in Eq.~\eqref{sec:superrl_loss} forces every instance to be
handled by \emph{either} supervised imitation or policy gradients.  We explore
two relaxations that keep the notation of
Sec.~\ref{sec:superrl_loss}—namely
\(A_k,\;r_k,\;g(\cdot,\cdot)\)—while softening the coupling between the two
learning signals.

\paragraph{Advantage-based soft gate.}
We define an alternative binary gate that triggers RL only if \emph{at least one} trajectory has a strictly positive advantage:
\[
c_{\!\mathcal{A}}(x)=\mathbf{1}\!\bigl(\max_k A_k>0\bigr)
\qquad\text{(triggered if any rollout outperforms the baseline)}
\]

The resulting loss is:
\[
\boxed{\;
\mathcal{L}_{\text{AdvGate}}(\theta;x)
  =(1-c_{\!\mathcal{A}}(x))\,\mathcal{L}_{\mathrm{SFT}}(\theta;x)
  +      c_{\!\mathcal{A}}(x)\,\mathcal{L}_{\mathrm{PG}}(\theta;x)
\;}
\tag{1}
\]
Unlike the original \textsc{SuperRL} switch, which triggers RL updates whenever any rollout yields a nonzero \emph{raw} reward, the advantage-based gate examines whether any trajectory yields a strictly \emph{positive advantage}—that is, outperforms a learned baseline \(b(x)\). This makes it more selective: dense but uninformative reward patterns (e.g., constant shaping terms) are often cancelled out by the baseline, resulting in \(A_k = 0\) and thus closing the gate. Even when raw rewards are nonzero, the update may fall back to SFT if no genuine improvement is detected. This design prevents futile RL updates on deceptively positive but sub-optimal trajectories and is particularly effective in sparse-reward settings with frequent shaping noise or misleading reward signals.

\paragraph{Uncertainty-weighted hybrid fusion.}
A fully continuous alternative removes the binary gate and learns how much to
trust each loss through two log-variance scalars
\(\sigma_{\mathrm{pg}}\) and \(\sigma_{\mathrm{sft}}\):
\[
\boxed{
  \mathcal{L}_{\text{Hybrid}}(\theta)=
     e^{-2\sigma_{\mathrm{pg}}}\,\mathcal{L}_{\mathrm{PG}}(\theta)
     +e^{-2\sigma_{\mathrm{sft}}}\,\mathcal{L}_{\mathrm{SFT}}(\theta)
     +\sigma_{\mathrm{pg}}+\sigma_{\mathrm{sft}}
}\tag{2}
\]
where \(\mathcal{L}_{\mathrm{PG}}\) may be instantiated by
PPO, TRPO, or GRPO.  Demonstration and on-policy batches are mixed
within every update; the learned variances down-weight noisy gradients yet the
additive \(\sigma\)-terms stop either branch from collapsing to zero.
During training, we interleave high-quality demonstrations with roll-outs in
each mini-batch.  The two uncertainty parameters adaptively balance signal
quality without any explicit switching rule: noisy or unstable gradients are
automatically suppressed by the factors \(e^{-2\sigma}\), while the additive
\(\sigma\) penalties prevent the model from becoming overconfident.

Equation~(1) preserves the hyperparameter-free spirit of the original gate
while being more selective; Eq.~(2) offers a differentiable, learnable fusion
at the cost of two extra parameters.

\section{Experiments}
\subsection{Experimental Setup}

We evaluate \method across a diverse suite of reasoning benchmarks that vary in structure, difficulty, and reward sparsity. Our evaluation spans both \textit{dense-reward tasks} (e.g., \textsc{GSM8K}, \textsc{MetaMath}) and \textit{sparse-reward tasks} (e.g., \textsc{OpenR1}, \textsc{PRM12K}). In all settings, \method consistently outperforms strong baselines, including pure RL and the conventional two-stage SFT+RL pipeline.

In dense regimes, \method performs comparably or better than RL, suggesting that the SFT fallback is rarely triggered and thus incurs no degradation. In sparse regimes, where rollouts frequently fail, \method activates its supervised fallback, leading to significant gains in success rate, generalization, and training stability.

To gain deeper insights into training dynamics, we track the proportion of samples that trigger RL versus SFT over time. On sparse tasks, SFT is used frequently at the beginning to bootstrap learning and is phased out as the policy improves. In contrast, dense tasks rarely invoke SFT. This per-instance adaptivity balances exploration and supervision effectively, reducing KL divergence variance and promoting stable optimization.

We further ablate several design variants: (1) \textit{Advantage-based gating}, which replaces reward-checking with advantage-checking; (2) \textit{Hybrid-Log-Sigma}, which softly blends RL and SFT losses using learned uncertainty weights; and (3) \textit{Fixed-schedule hybrids} with manually tuned schedules. Although these variants yield some improvements over naïve RL, they require extra tuning or introduce rigidity. In contrast, \method offers a simple, hyperparameter-free, and plug-and-play switching mechanism that delivers stronger and more stable performance.

\begin{table*}[ht]
\centering
\caption{
Cross-dataset generalization accuracy (\%) of different training methods. Each row block corresponds to a model trained on a specific dataset (e.g., GSM8K, Metamath, etc.), while columns report the zero-shot accuracy on various held-out test sets. The baseline (\emph{Vanilla RL}) performance is shown in the first row of each block and shaded in gray. Colored cells indicate the relative change compared to this baseline: \textcolor{green}{green} for improvement and \textcolor{red}{red} for degradation, with darker shades representing larger absolute differences. This visualization highlights the generalization patterns of each training method across diverse reasoning tasks.
}
\resizebox{\textwidth}{!}{%
\begin{tabular}{llccccccc}
\toprule
\textbf{Training} & \textbf{Method} & \textbf{GSM8K} & \textbf{Metamath} & \textbf{PRM12K} & \textbf{LIMO} & \textbf{OpenR1} & \textbf{AIME24} & \textbf{AIME25} \\
\midrule
\multirow{4}{*}{GSM8K}
 & RL      & \cellcolor{gray!15}72.29 & \cellcolor{gray!15}72.05 & \cellcolor{gray!15}33.18 & \cellcolor{gray!15}23.58 & \cellcolor{gray!15}14.96 & \cellcolor{gray!15}2.92 & \cellcolor{gray!15}1.10 \\
 & SFT             & \cellcolor{red!48}58.00 & \cellcolor{red!30}64.29 & \cellcolor{red!45}22.86 & \cellcolor{red!60}0.00  & \cellcolor{red!30}7.14  & \cellcolor{red!15}0.00 & \cellcolor{red!15}0.00 \\
 & SFT+RL          & \cellcolor{red!5}71.45  & \cellcolor{green!15}74.56 & \cellcolor{red!5}32.14 & \cellcolor{red!30}18.38 & \cellcolor{red!5}14.58 & \cellcolor{red!15}1.45 & \cellcolor{red!5}0.44 \\
 & \textbf{SuperRL}& \cellcolor{green!30}78.86 & \cellcolor{green!30}77.02 & \cellcolor{green!30}36.96 & \cellcolor{green!60}53.46 & \cellcolor{green!15}16.11 & \cellcolor{green!15}4.10 & \cellcolor{green!15}1.81 \\
\midrule
\multirow{4}{*}{Metamath}
 & RL      & \cellcolor{gray!15}76.21 & \cellcolor{gray!15}81.03 & \cellcolor{gray!15}38.16 & \cellcolor{gray!15}41.63 & \cellcolor{gray!15}15.01 & \cellcolor{gray!15}6.67 & \cellcolor{gray!15}2.05 \\
 & SFT             & \cellcolor{red!30}69.43 & \cellcolor{red!30}74.29 & \cellcolor{red!60}22.14 & \cellcolor{red!60}14.29 & \cellcolor{red!15}12.86 & \cellcolor{red!30}0.00 & \cellcolor{green!15}3.33 \\
 & SFT+RL          & \cellcolor{red!15}71.44 & \cellcolor{red!30}72.14 & \cellcolor{red!30}32.14 & \cellcolor{red!60}9.74  & \cellcolor{red!15}12.31 & \cellcolor{red!15}2.99 & \cellcolor{red!15}0.64 \\
 & \textbf{SuperRL}& \cellcolor{green!8}77.00 & \cellcolor{green!15}84.29 & \cellcolor{green!15}39.09 & \cellcolor{green!30}48.08 & \cellcolor{green!8}15.73 & \cellcolor{red!15}4.73 & \cellcolor{red!5}1.81 \\
\midrule
\multirow{4}{*}{LIMO}
 & RL      & \cellcolor{gray!15}73.28 & \cellcolor{gray!15}76.66 & \cellcolor{gray!15}37.26 & \cellcolor{gray!15}44.13 & \cellcolor{gray!15}16.82 & \cellcolor{gray!15}3.79 & \cellcolor{gray!15}1.66 \\
 & SFT             & \cellcolor{red!60}39.14 & \cellcolor{red!60}35.00 & \cellcolor{red!60}21.43 & \cellcolor{red!60}0.00  & \cellcolor{red!30}10.71 & \cellcolor{red!15}0.00 & \cellcolor{green!15}3.33 \\
 & SFT+RL          & \cellcolor{green!15}77.44 & \cellcolor{green!5}76.72 & \cellcolor{red!5}35.75 & \cellcolor{red!20}34.88 & \cellcolor{red!5}15.24 & \cellcolor{red!5}2.85 & \cellcolor{green!30}6.67 \\
 & \textbf{SuperRL}& \cellcolor{gray!15}73.28 & \cellcolor{red!15}66.43 & \cellcolor{red!5}35.64 & \cellcolor{red!15}36.99 & \cellcolor{red!5}14.92 & \cellcolor{green!5}3.99 & \cellcolor{red!5}1.14 \\
\midrule
\multirow{4}{*}{PRM12K}
 & RL      & \cellcolor{gray!15}64.61 & \cellcolor{gray!15}74.40 & \cellcolor{gray!15}44.48 & \cellcolor{gray!15}25.14 & \cellcolor{gray!15}19.65 & \cellcolor{gray!15}2.59 & \cellcolor{gray!15}2.09 \\
 & SFT             & \cellcolor{red!15}61.43 & \cellcolor{red!60}57.86 & \cellcolor{red!60}22.86 & \cellcolor{red!60}0.00  & \cellcolor{red!45}11.43 & \cellcolor{red!15}0.00 & \cellcolor{red!15}0.00 \\
 & SFT+RL          & \cellcolor{green!15}67.56 & \cellcolor{green!30}80.46 & \cellcolor{red!5}42.99 & \cellcolor{green!15}28.36 & \cellcolor{red!15}16.61 & \cellcolor{green!5}2.71 & \cellcolor{red!5}1.21 \\
 & \textbf{SuperRL}& \cellcolor{green!30}70.95 & \cellcolor{green!30}81.45 & \cellcolor{green!15}48.10 & \cellcolor{green!48}39.57 & \cellcolor{red!5}19.35 & \cellcolor{green!15}4.28 & \cellcolor{green!5}2.51 \\
\midrule
\multirow{4}{*}{OpenR1}
 & RL      & \cellcolor{gray!15}56.98 & \cellcolor{gray!15}67.78 & \cellcolor{gray!15}42.10 & \cellcolor{gray!15}37.42 & \cellcolor{gray!15}17.40 & \cellcolor{gray!15}2.91 & \cellcolor{gray!15}0.99 \\
 & SFT             & \cellcolor{green!48}65.00 & \cellcolor{red!60}40.71 & \cellcolor{red!48}28.57 & \cellcolor{red!30}28.57 & \cellcolor{red!30}11.43 & \cellcolor{green!5}3.33 & \cellcolor{red!5}0.00 \\
 & SFT+RL          & \cellcolor{green!48}66.77 & \cellcolor{red!5}67.53 & \cellcolor{red!45}30.66 & \cellcolor{red!45}26.98 & \cellcolor{red!15}14.93 & \cellcolor{green!5}3.04 & \cellcolor{green!5}1.42 \\
 & \textbf{SuperRL}& \cellcolor{green!60}73.68 & \cellcolor{green!48}77.98 & \cellcolor{red!15}37.21 & \cellcolor{red!15}33.62 & \cellcolor{green!15}19.68 & \cellcolor{green!5}3.55 & \cellcolor{green!5}1.84 \\
\bottomrule
\end{tabular}
}
\label{tab:cross_dataset_colored}
\end{table*}

\subsection{Cross-Dataset Generalization Analysis}
\label{sec:cross_dataset}

\begin{table}[ht]
\centering
\caption{Cross-dataset test accuracy (\%) of models trained on \textbf{Hitab} dataset. Colors indicate change relative to \emph{RL} (top row): green = higher accuracy, red = lower accuracy; deeper shades reflect larger deltas.}
\resizebox{\textwidth}{!}{%
\begin{tabular}{llcccccccc}
\toprule
\textbf{Training} & \textbf{Method} & \textbf{GSM8K} & \textbf{Meta\-math} & \textbf{PRM12K} & \textbf{LIMO} & \textbf{OpenR1} & \textbf{AIME24} & \textbf{AIME25} & \textbf{Hitab} \\
\midrule
\multirow{4}{*}{Hitab}
 & RL      & \cellcolor{gray!15}72.05 & \cellcolor{gray!15}44.21 & \cellcolor{gray!15}26.63 & \cellcolor{gray!15}10.94 & \cellcolor{gray!15}15.26 & \cellcolor{gray!15}2.99 & \cellcolor{gray!15}0.57 & \cellcolor{gray!15}30.03 \\
 & SFT             & \cellcolor{red!30}64.29  & \cellcolor{green!30}55.00 & \cellcolor{green!30}33.57 & \cellcolor{green!48}28.57 & \cellcolor{red!15}11.43 & \cellcolor{red!15}0.00 & \cellcolor{red!5}0.00  & \cellcolor{green!15}33.33 \\
 & SFT+RL          & \cellcolor{green!15}74.56 & \cellcolor{green!60}72.53 & \cellcolor{green!30}35.91 & \cellcolor{green!60}36.93 & \cellcolor{green!10}16.65 & \cellcolor{green!10}4.41 & \cellcolor{green!15}4.11 & \cellcolor{green!30}38.64 \\
 & \textbf{SuperRL}& \cellcolor{green!30}77.40 & \cellcolor{red!30}34.29 & \cellcolor{green!15}29.29 & \cellcolor{green!30}19.43 & \cellcolor{red!15}12.45 & \cellcolor{green!8}3.33 & \cellcolor{green!15}2.08 & \cellcolor{red!48}18.47 \\
\bottomrule
\end{tabular}%
}
\label{tab:hitab_colored}
\end{table}

\paragraph{Key Observations}
Across the $5{\times}7=35$ pairs in Table~\ref{tab:cross_dataset_colored},
\textbf{SuperRL} improves upon \emph{RL} in \textbf{24} cases,
is worse in 10, and ties once.\footnote{%
  A cell is counted as an improvement when the SuperRL entry strictly exceeds the
  corresponding RL entry.}
Gains are largest when the \emph{source} dataset itself supplies either
dense rewards (e.g., \textsc{GSM8K}, \textsc{MetaMath})
or moderately sparse but still informative feedback (\textsc{PRM12K}, \textsc{OpenR1}). In these settings, reward–gated switching allows exploration‐driven RL updates while still injecting high-quality supervision on difficult examples, yielding improvements of up to \textbf{+29.9 pp}.

\paragraph{Insights from \textsc{LIMO}}
When trained on \textsc{LIMO}, SuperRL underperforms RL—a divergence attributable to both the dataset’s limited size and its deliberate construction. Specifically, LIMO contains only 817 examples, all manually filtered to emphasize high reasoning complexity. This results in sparse rewards and low pattern diversity: successful trajectories are few and narrowly distributed. In such settings, RL can overfit to these recurring patterns, achieving higher in-domain accuracy. However, SuperRL’s fallback to SFT introduces broader—but diffuse—supervision that struggles to provide useful learning signals when data is both scarce and hard. Consequently, SuperRL cannot effectively learn from LIMO alone. Nevertheless, SuperRL demonstrates strong generalization despite its poor in-domain performance. When trained on more diverse and densely supervised datasets such as \textsc{GSM8K}, it achieves surprisingly strong results on LIMO and other hard transfer benchmarks. This contrast highlights the critical role of training data: it not only affirms the limitations of LIMO's narrow construction but also reinforces SuperRL’s ability to generalize when given sufficient and well-distributed supervision. The findings suggest that the model’s underperformance is more reflective of dataset-specific bottlenecks than of architectural limitations.

Results in Table~\ref{tab:hitab_colored} show a clear contrast with earlier trends: the two-stage \textit{SFT+RL} achieves the best performance on \textsc{Hitab}, while SuperRL falls behind even vanilla RL. \textsc{Hitab} is a challenging, cross-domain benchmark focused on table reasoning, where rewards are binary and high-quality demonstrations—collected via DeepSeek R1 distillation and RFT-style sampling—are abundant. In this setting, a dedicated SFT phase proves critical: it allows the model to absorb diverse, domain-specific reasoning patterns before engaging in exploration. SuperRL, by contrast, switches prematurely to RL as soon as a single successful trajectory is found, which may truncate the supervision phase too early and miss the chance to learn stable table reasoning heuristics. This issue is especially pronounced in cross-domain settings like \textsc{Hitab}, where general reasoning strategies learned via minimal supervision may not transfer well. Nevertheless, SuperRL still exhibits strong generalization ability across datasets. Despite its underperformance on \textsc{Hitab}, it maintains competitive transfer performance on other challenging benchmarks—highlighting that its fallback supervision discourages brittle memorization and fosters more broadly applicable reasoning strategies.

\paragraph{Practical take-aways.}
\begin{itemize}\setlength{\itemsep}{2pt}
  \item \textbf{Reward-gated switching is generally effective}: it provides
        a plug-and-play improvement over RL with no extra hyperparameters
        in most transfer scenarios.
  \item \textbf{Dataset properties matter}: highly filtered, ultra-sparse
        sources such as LIMO may require additional stabilisation;  
        heterogeneous, demonstration-rich benchmarks such as Hitab benefit
        from longer SFT exposure.
  \item \textbf{Future work}: learn a data-driven gating schedule that
        adapts to both reward sparsity and demonstration density, thereby
        unifying the advantages of SuperRL and two-stage SFT+RL.
\end{itemize}

\subsection{Adaptive Behaviors During Training}
\label{sec:adaptive_behaviors_during_training}
Beyond accuracy and generalization, we also assess \method in terms of training stability and computational cost. These dimensions are critical for real-world deployment, especially in settings where reward sparsity and rollout variance can induce unstable gradients and excessive tuning overhead.

\paragraph{Entropy Stabilization Patterns.}
\begin{table}[t]
\centering
\caption{Entropy statistics of RL and SuperRL across datasets. Green cells indicate improvements over RL (i.e., lower entropy), with darker shades representing greater reductions.}
\label{tab:entropy_stats}
\begin{tabular}{lcccc}
\toprule
\textbf{Method \& Dataset} & \textbf{Max} & \textbf{Min} & \textbf{Range} & \textbf{Variance} \\
\midrule
\rowcolor{gray!10}
RL (LIMO)          & 5.794 & 0.006 & 5.788 & 0.831 \\
SuperRL (LIMO)     & \cellcolor{green!30}4.698 & \cellcolor{gray!0}0.009 & \cellcolor{green!30}4.689 & \cellcolor{green!50}0.490 \\
\rowcolor{gray!10}
RL (GSM8K)         & 4.850 & 0.004 & 4.846 & 0.171 \\
SuperRL (GSM8K)    & \cellcolor{gray!0}4.875 & \cellcolor{green!10}0.006 & \cellcolor{gray!0}4.869 & \cellcolor{green!20}0.134 \\
\rowcolor{gray!10}
RL (Metamath)      & 4.863 & 0.016 & 4.847 & 0.424 \\
SuperRL (Metamath) & \cellcolor{gray!0}5.417 & \cellcolor{gray!0}0.020 & \cellcolor{gray!0}5.397 & \cellcolor{green!10}0.405 \\
\rowcolor{gray!10}
RL (OpenR1)        & 4.978 & 0.044 & 4.934 & 0.708 \\
SuperRL (OpenR1)   & \cellcolor{gray!0}5.208 & \cellcolor{green!10}0.034 & \cellcolor{gray!0}5.174 & \cellcolor{green!10}0.695 \\
\rowcolor{gray!10}
RL (HITAB)         & 6.002 & 0.000 & 6.002 & 0.363 \\
SuperRL (HITAB)    & \cellcolor{green!40}5.376 & \cellcolor{gray!0}0.006 & \cellcolor{green!40}5.370 & \cellcolor{green!50}0.186 \\
\bottomrule
\end{tabular}
\end{table}
To evaluate training stability, we analyze the entropy statistics of RL and \method across multiple datasets, as shown in Table~\ref{tab:entropy_stats}. Entropy quantifies the uncertainty of the policy during exploration; high variance or wide ranges often indicate unstable updates, while excessively low entropy can suppress exploration. A well-calibrated entropy profile reflects both stable learning and sufficient exploration capacity.

Across most datasets, \method achieves a meaningful reduction in entropy variance and range compared to vanilla RL, suggesting more stable and controlled optimization. For instance, in \textsc{GSM8K} and \textsc{Metamath}, \method consistently narrows the entropy range and lowers the variance, leading to smoother convergence without sacrificing exploration. These improvements reflect the benefit of selectively falling back to supervised fine-tuning in failure cases.

However, on \textsc{LIMO} and \textsc{HITAB}, we observe an over-suppression of entropy dynamics: while \method lowers the entropy variance (e.g., from $0.831$ to $0.490$ on \textsc{LIMO}), it also curtails the range and maximum entropy to a degree that may hinder sufficient exploration. This suggests that frequent fallback to SFT may anchor the policy too tightly to demonstrative behaviors, limiting the diversity of trajectories explored. Consequently, the model may prematurely exploit suboptimal behaviors, leading to diminished performance on exploration-heavy tasks.

These findings highlight that entropy stabilization is a double-edged sword: while reducing variance helps avoid erratic learning, overly dampened entropy signals can impair policy expressiveness. \method performs best when the fallback mechanism maintains a dynamic balance between policy stability and exploratory diversity. In aggregate, its adaptive optimization strategy yields more reliable convergence across diverse reward landscapes, though careful tuning may be necessary to avoid under-exploration in sparse regimes.

\paragraph{SFT Trigger Patterns.}
\begin{figure}[t]
\centering
\includegraphics[width=0.95\linewidth]{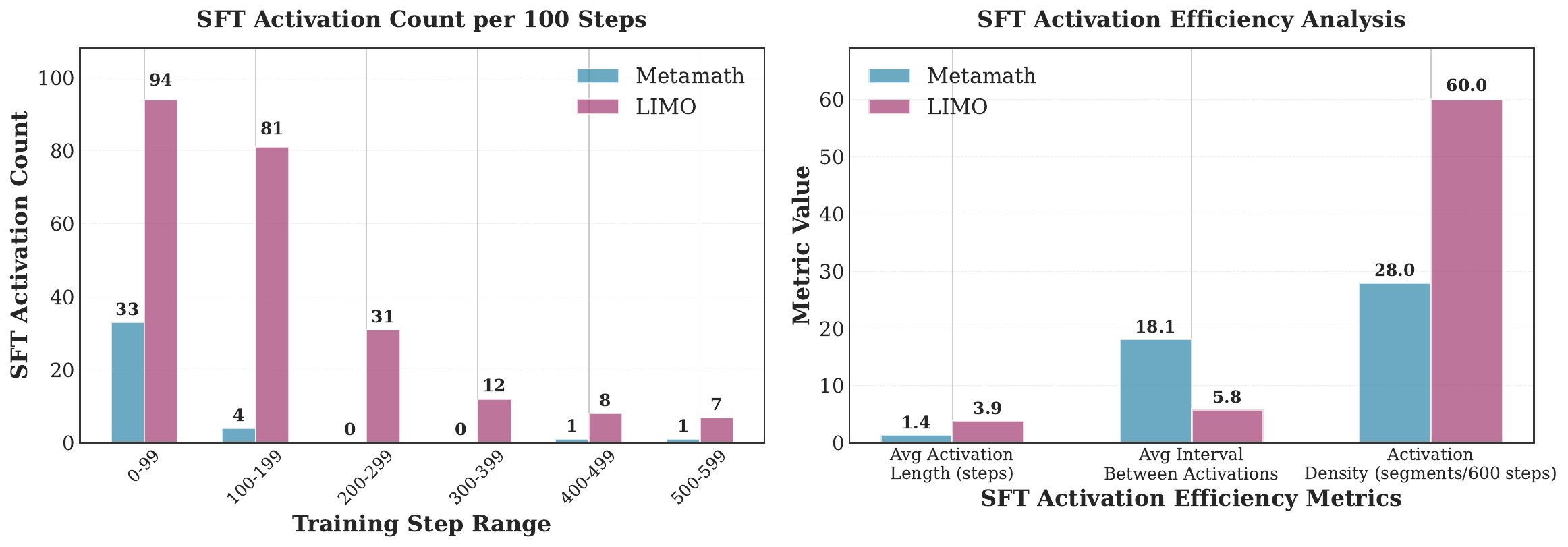}
\caption{
\textbf{Analysis of SFT Activation Patterns and Efficiency.}
\textbf{Left:} The histogram shows the count of supervised fine-tuning (SFT) activations per 100 training steps for two datasets: \textit{Metamath} and \textit{LIMO}. Most SFT activations in both cases occur in the early phase of training (0--100 steps), with \textit{LIMO} exhibiting a denser activation pattern overall.
\textbf{Right:} A comparison of SFT activation efficiency metrics across datasets. \textit{LIMO} shows longer average activation lengths, shorter intervals between activations, and a higher activation density (segments with SFT triggered per 500 steps), indicating more frequent and sustained reliance on SFT throughout training.
}
\label{fig:sft_trigger_comparison}
\vspace{0.5em}
\end{figure}

We observe notable differences in how frequently and efficiently supervised fine-tuning (SFT) is triggered across datasets. As shown in Figure~\ref{fig:sft_trigger_comparison}, \textit{LIMO} triggers SFT much more frequently than \textit{Metamath}, especially in the early stages of training (0–100 steps), suggesting that it encounters a higher proportion of hard instances requiring fallback supervision. Additionally, the right panel highlights that \textit{LIMO} exhibits longer average SFT activation lengths and denser activation patterns, with shorter intervals between activations. These trends imply that the model struggles to sustain reward-guided learning for longer stretches and must rely more consistently on offline signals. In contrast, \textit{Metamath} shows more sparsely distributed SFT usage, indicating more stable or successful online exploration with less dependence on supervised corrections once initial learning has occurred.

Importantly, across both datasets, we observe a consistent trend: the reliance on SFT signals decreases as training progresses. This pattern highlights the adaptive nature of our method—SFT is heavily used in early stages to guide learning when the model lacks competency, but gradually fades out as the policy improves and the model learns to solve problems correctly on its own. This transition reflects a shift from supervision-driven learning to reward-driven generalization, demonstrating the effectiveness of our hybrid training framework.

Table~\ref{tab:sft_score_stats} provides a quantitative summary of average scores and SFT fallback usage across datasets. A clear negative correlation emerges between the SuperRL average score and total SFT trigger count ($r = -0.86$, $p = 0.028\ <\ 0.05$), suggesting that fallback is most critical on tasks where standard RL struggles to achieve high scores. This trend is even more pronounced when comparing score gains: in datasets with high SFT fallback (e.g., \textsc{PRM12K}, \textsc{OpenR1}), SuperRL significantly outperforms RL. Conversely, in easier datasets like \textsc{GSM8K} or \textsc{Metamath}, where scores are relatively high and SFT is rarely needed, the margin is small or even reversed.

This analysis underscores the adaptive value of SuperRL's fallback mechanism. When RL alone fails to make progress—either due to sparse rewards or brittle dynamics—SFT provides essential guidance. Rather than rely on static supervision, SuperRL selectively triggers imitation based on rollout outcomes, offering a robust and flexible solution for navigating varying task difficulty. The heterogeneous SFT usage patterns thus reflect not only task complexity but also the system’s capacity to maintain learning momentum across regimes.

Interestingly, when analyzing the relationship between RL average score and SFT trigger count, we observe a similar but weaker negative correlation ($r = -0.64$, $p \approx 0.171$). While the trend aligns with the intuition that lower RL performance invites more SFT fallback, the lack of statistical significance suggests that raw RL scores alone are not as strong a predictor of fallback reliance. This is likely because SuperRL selectively invokes SFT based on rollout success rather than static task characteristics—meaning that even if RL achieves moderately good average scores on some tasks, it may still benefit from occasional fallback in rare failure cases.

In contrast, the SuperRL score shows a much stronger and statistically significant correlation with SFT usage, reinforcing the idea that SuperRL dynamically calibrates its reliance on supervision to support progress where RL plateaus. Taken together, these results highlight that SFT fallback not only enhances performance but also reflects the model’s internal learning difficulty, offering a principled mechanism to bridge the gap between exploration and convergence.

\begin{table}[t]
\centering
\caption{Average score and SFT-triggered update count during training. Higher SFT usage correlates with harder tasks and greater gains in average score.}
\label{tab:sft_score_stats}
\begin{tabular}{lccc}
\toprule
\textbf{Dataset} & \textbf{RL Score} & \textbf{SuperRL Score} & \textbf{SFT Trigger Count} \\
\midrule
OpenR1    & 0.526 & 0.619 & 728 \\
PRM12K    & 0.268 & 0.590 & 484 \\
HITAB     & 0.645 & 0.665 & 270 \\
LIMO      & 0.500 & 0.688 & 247 \\
Metamath  & 0.715 & 0.700 & 40 \\
GSM8K     & 0.687 & 0.758 & 19 \\
\bottomrule
\end{tabular}
\end{table}

\subsection{Hybrid Variants and Insights}

We conclude by comparing \method with two hybrid variants\ —\ \textit{Hybrid-Advantage-Gated} and \textit{Hybrid-Log-Sigma}, each relaxing the original reward-gated switch in different ways. Table~\ref{tab:hybrid_variant_table} presents the cross-dataset generalization performance under consistent training datasets.

\begin{table}[ht]
\centering
\small            
\setlength{\tabcolsep}{3.4pt}
\renewcommand{\arraystretch}{1.1}
\begin{tabular}{@{}lcccccccc@{}}
\toprule
\textbf{Training Set} & \textbf{Method} & \textbf{GSM8K} & \textbf{Metamath} & \textbf{PRM12K} & \textbf{LIMO} & \textbf{OpenR1} & \textbf{AIME2024} & \textbf{AIME2025} \\
\midrule
\multirow{3}{*}{GSM8K} 
    & SuperRL   & 78.86 & 77.02 & 36.96 & 53.46 & 16.11 & 4.10 & 1.81 \\
    & Adv-Gated & \textcolor{red}{76.43} & \textcolor{green}{77.93} & \textcolor{green}{37.48} & \textcolor{red}{34.73} & \textcolor{red}{15.50} & \textcolor{red}{3.46} & \textcolor{green}{2.20} \\
    & Log-Sigma & \textcolor{red}{73.88} & \textcolor{red}{52.72} & \textcolor{green}{37.60} & \textcolor{red}{24.18} & \textcolor{red}{14.29} & \textcolor{red}{2.95} & \textcolor{red}{1.22} \\
\cmidrule(lr){1-9}  
\multirow{3}{*}{Metamath} 
    & SuperRL   & 77.00 & 84.29 & 39.09 & 48.08 & 15.73 & 4.73 & 1.81 \\
    & Adv-Gated & \textcolor{red}{74.77} & 84.29 & \textcolor{red}{38.26} & \textcolor{red}{43.27} & \textcolor{green}{15.86} & \textcolor{red}{4.21} & \textcolor{green}{2.87} \\
    & Log-Sigma & \textcolor{red}{71.95} & \textcolor{red}{77.71} & \textcolor{red}{32.60} & \textcolor{red}{14.80} & \textcolor{red}{14.21} & \textcolor{red}{2.50} & \textcolor{red}{1.02} \\
\cmidrule(lr){1-9}
\multirow{3}{*}{LIMO} 
    & SuperRL   & 73.28 & 66.43 & 35.64 & 36.99 & 14.92 & 3.99 & 1.14 \\
    & Adv-Gated & \textcolor{red}{71.20} & \textcolor{green}{77.39} & \textcolor{green}{36.98} & \textcolor{green}{37.54} & \textcolor{green}{15.24} & \textcolor{green}{4.63} & \textcolor{green}{6.67} \\
    & Log-Sigma & \textcolor{red}{70.55} & \textcolor{green}{71.86} & \textcolor{red}{34.21} & \textcolor{red}{28.96} & \textcolor{red}{14.66} & \textcolor{red}{3.63} & \textcolor{green}{2.36} \\
\cmidrule(lr){1-9}
\multirow{3}{*}{PRM12K} 
    & SuperRL   & 70.95 & 81.45 & 48.10 & 39.57 & 19.35 & 4.28 & 2.51 \\
    & Adv-Gated & \textcolor{green}{74.90} & \textcolor{red}{81.18} & \textcolor{green}{51.00} & \textcolor{green}{40.87} & \textcolor{green}{20.17} & \textcolor{red}{2.26} & \textcolor{red}{2.11} \\
    & Log-Sigma & \textcolor{red}{60.97} & \textcolor{red}{46.98} & \textcolor{red}{42.94} & \textcolor{red}{18.06} & \textcolor{red}{15.87} & \textcolor{red}{1.51} & \textcolor{red}{1.86} \\
\cmidrule(lr){1-9}
\multirow{3}{*}{OpenR1} 
    & SuperRL   & 73.68 & 77.98 & 37.21 & 33.62 & 19.68 & 3.55 & 1.84 \\
    & Adv-Gated & \textcolor{red}{14.28} & \textcolor{red}{29.61} & \textcolor{red}{28.57} & \textcolor{red}{14.11} & \textcolor{red}{10.23} & \textcolor{red}{0.56} & \textcolor{red}{0.74} \\
    & Log-Sigma & \textcolor{red}{10.40} & \textcolor{red}{27.18} & \textcolor{red}{27.53} & \textcolor{red}{14.29} & \textcolor{red}{12.18} & \textcolor{red}{0.18} & \textcolor{red}{1.42} \\
\cmidrule(lr){1-9}
\multirow{3}{*}{Hitab} 
    & SuperRL   & 77.40 & 34.29 & 29.29 & 19.43 & 12.45 & 3.33 & 2.08 \\
    & Adv-Gated & \textcolor{green}{78.09} & \textcolor{green}{38.62} & 29.29 & \textcolor{red}{9.76} & \textcolor{green}{13.51} & \textcolor{green}{5.54} & \textcolor{red}{0.64} \\
    & Log-Sigma & \textcolor{red}{61.08} & \textcolor{red}{27.48} & \textcolor{red}{29.02} & \textcolor{red}{15.78} & \textcolor{red}{12.18} & \textcolor{red}{3.05} & \textcolor{red}{1.09} \\
\bottomrule
\end{tabular}
\caption{Cross-dataset test accuracy (\%) of SuperRL and its variants. Adv-Gated uses advantage-based gating; Log-Sigma softly blends RL and SFT losses with uncertainty-weighted averaging. Color indicates delta relative to SuperRL: \textcolor{green}{green} = higher, \textcolor{red}{red} = lower.}
\label{tab:hybrid_variant_table}
\end{table}

\paragraph{SuperRL Outperforms in the Majority.}

Despite its localized shortcomings on \textsc{LIMO} and \textsc{Hitab}, \textbf{SuperRL consistently outperforms both Hybrid-Adv-Gated and Hybrid-Log-Sigma in the vast majority of settings}. As shown in Table~\ref{tab:hybrid_variant_table}, SuperRL achieves higher or equal accuracy in 28 out of 42 train-test pairs compared to the advantage-gated variant, and dominates the uncertainty-based hybrid in nearly every case. This pattern holds especially true when the training dataset offers a reasonable mix of reward sparsity and demonstration richness—conditions under which SuperRL's reward-gated fallback efficiently balances exploration and imitation.

These results reaffirm the core design philosophy behind SuperRL: by using a simple, deterministic switch based on actual rollout outcomes, it avoids overfitting to noisy shaping signals while also bypassing the need for complex auxiliary mechanisms. In contrast, advantage-based gating introduces extra learning dynamics that may require careful calibration, while soft blending with uncertainty weights often leads to instability or underperformance. The empirical gap across datasets highlights that SuperRL’s binary gating mechanism—though minimal—is highly effective and broadly applicable.

\paragraph{When SuperRL Falls Behind: The Case of \textsc{LIMO} and \textsc{Hitab}.}

While \method generally achieves strong generalization across diverse benchmarks, two notable exceptions emerge: \textsc{LIMO} and \textsc{Hitab}. In both cases, the \textit{Advantage-Gated} variant surpasses SuperRL, suggesting specific regimes where reward-gated switching may be less effective.

In \textsc{LIMO}, the dataset is both small in size (817 samples) and manually curated for high reasoning difficulty, resulting in extremely sparse rewards and low structural diversity. Under such conditions, SuperRL's reward-based gate may be too permissive—deeming trajectories with minimal reward as successful, thereby deactivating the fallback to supervision prematurely. The \textit{Advantage-Gated} method, which relies on learned advantage signals, is more selective: it delays the switch until a trajectory meaningfully improves over the baseline. This added caution helps preserve supervision longer during early training, which proves crucial when reward signals are misleading or insufficient for exploration.

In \textsc{Hitab}, the situation is different but equally revealing. Hitab focuses on table-based reasoning, which presents a different set of challenges compared to math word problems or symbolic reasoning. Furthermore, the dataset is constructed via distillation from DeepSeek R1 followed by RFT-style filtering, resulting in trajectories that reflect model-like patterns rather than diverse human logic. In this context, SuperRL’s reward-triggered switching can be overly aggressive: once it detects any success signal—even if superficial—it disables the SFT path, missing the opportunity to deeply absorb structured table reasoning heuristics. On the other hand, the \textit{Advantage-Gated} approach defers the switch until the learned advantage becomes significant, thus sustaining supervision longer and capturing the domain-specific inductive biases required for table reasoning.

\paragraph{Interpretation and Value of Reward-Gated Switching.}

These failure cases highlight an important limitation: although reward-based gating is efficient and principled in most reasoning tasks, it may struggle when (i) reward signals are noisy and overly sparse (e.g., \textsc{LIMO}), or (ii) task structure demands more sustained exposure to supervision (e.g., table-based reasoning in \textsc{Hitab}).

Nevertheless, SuperRL remains highly competitive overall. Its performance on other hard benchmarks—especially when trained on more general-purpose datasets like \textsc{GSM8K} or \textsc{PRM12K}—shows that the reward-gated fallback mechanism is effective in dynamically routing learning signals based on rollout success, without the need for extra models or hyperparameters.

\paragraph{Takeaway.}
While \textit{Advantage-Gated} may outperform in specific scenarios like \textsc{LIMO} and \textsc{Hitab}, SuperRL offers a broadly applicable, plug-and-play strategy that delivers reliable performance across varied reasoning domains. Future work may explore combining reward- and advantage-based signals or developing more domain-aware switching strategies—but the current design already strikes a strong balance between stability, adaptability, and simplicity.

\section{Conclusion}
We present \method, a unified training framework that adaptively combines supervised and reinforcement signals to improve reasoning under both dense and sparse rewards. Experiments show that SuperRL achieves superior performance, stability, and generalization across diverse reasoning benchmarks. Limitations and future work are discussed in Appendix~\ref{sec:limitation}.

\newpage



\bibliography{references}

\begin{thebibliography}{10}

\bibitem{wei2022chain}
Jason Wei, Xuezhi Wang, Dale Schuurmans, Maarten Bosma, Fei Xia, Ed~Chi, Quoc~V Le, Denny Zhou, et~al.
\newblock Chain-of-thought prompting elicits reasoning in large language models.
\newblock {\em Advances in neural information processing systems}, 35:24824--24837, 2022.

\bibitem{wang2022self}
Xuezhi Wang, Jason Wei, Dale Schuurmans, Quoc Le, Ed~Chi, Sharan Narang, Aakanksha Chowdhery, and Denny Zhou.
\newblock Self-consistency improves chain of thought reasoning in language models.
\newblock {\em arXiv preprint arXiv:2203.11171}, 2022.

\bibitem{brown2020language}
Tom Brown, Benjamin Mann, Nick Ryder, Melanie Subbiah, Jared~D Kaplan, Prafulla Dhariwal, Arvind Neelakantan, Pranav Shyam, Girish Sastry, Amanda Askell, et~al.
\newblock Language models are few-shot learners.
\newblock {\em Advances in neural information processing systems}, 33:1877--1901, 2020.

\bibitem{wang2022towards}
Boshi Wang, Sewon Min, Xiang Deng, Jiaming Shen, You Wu, Luke Zettlemoyer, and Huan Sun.
\newblock Towards understanding chain-of-thought prompting: An empirical study of what matters.
\newblock {\em arXiv preprint arXiv:2212.10001}, 2022.

\bibitem{zhang2022automatic}
Zhuosheng Zhang, Aston Zhang, Mu~Li, and Alex Smola.
\newblock Automatic chain of thought prompting in large language models.
\newblock {\em arXiv preprint arXiv:2210.03493}, 2022.

\bibitem{cao2023graphreason}
Lang Cao.
\newblock Graphreason: Enhancing reasoning capabilities of large language models through a graph-based verification approach.
\newblock {\em arXiv preprint arXiv:2308.09267}, 2023.

\bibitem{zhang2025surveytesttimescalinglarge}
Qiyuan Zhang, Fuyuan Lyu, Zexu Sun, Lei Wang, Weixu Zhang, Wenyue Hua, Haolun Wu, Zhihan Guo, Yufei Wang, Niklas Muennighoff, Irwin King, Xue Liu, and Chen Ma.
\newblock A survey on test-time scaling in large language models: What, how, where, and how well?, 2025.

\bibitem{shao2024deepseekmath}
Zhihong Shao, Peiyi Wang, Qihao Zhu, Runxin Xu, Junxiao Song, Xiao Bi, Haowei Zhang, Mingchuan Zhang, YK~Li, Y~Wu, et~al.
\newblock Deepseekmath: Pushing the limits of mathematical reasoning in open language models.
\newblock {\em arXiv preprint arXiv:2402.03300}, 2024.

\bibitem{deepseekai2025deepseekr1incentivizingreasoningcapability}
DeepSeek-AI, Daya Guo, Dejian Yang, Haowei Zhang, Junxiao Song, Ruoyu Zhang, Runxin Xu, Qihao Zhu, Shirong Ma, Peiyi Wang, Xiao Bi, Xiaokang Zhang, Xingkai Yu, Yu~Wu, Z.~F. Wu, Zhibin Gou, Zhihong Shao, Zhuoshu Li, Ziyi Gao, Aixin Liu, Bing Xue, Bingxuan Wang, Bochao Wu, Bei Feng, Chengda Lu, Chenggang Zhao, Chengqi Deng, Chenyu Zhang, Chong Ruan, Damai Dai, Deli Chen, Dongjie Ji, Erhang Li, Fangyun Lin, Fucong Dai, Fuli Luo, Guangbo Hao, Guanting Chen, Guowei Li, H.~Zhang, Han Bao, Hanwei Xu, Haocheng Wang, Honghui Ding, Huajian Xin, Huazuo Gao, Hui Qu, Hui Li, Jianzhong Guo, Jiashi Li, Jiawei Wang, Jingchang Chen, Jingyang Yuan, Junjie Qiu, Junlong Li, J.~L. Cai, Jiaqi Ni, Jian Liang, Jin Chen, Kai Dong, Kai Hu, Kaige Gao, Kang Guan, Kexin Huang, Kuai Yu, Lean Wang, Lecong Zhang, Liang Zhao, Litong Wang, Liyue Zhang, Lei Xu, Leyi Xia, Mingchuan Zhang, Minghua Zhang, Minghui Tang, Meng Li, Miaojun Wang, Mingming Li, Ning Tian, Panpan Huang, Peng Zhang, Qiancheng Wang, Qinyu Chen, Qiushi Du, Ruiqi Ge, Ruisong
  Zhang, Ruizhe Pan, Runji Wang, R.~J. Chen, R.~L. Jin, Ruyi Chen, Shanghao Lu, Shangyan Zhou, Shanhuang Chen, Shengfeng Ye, Shiyu Wang, Shuiping Yu, Shunfeng Zhou, Shuting Pan, S.~S. Li, Shuang Zhou, Shaoqing Wu, Shengfeng Ye, Tao Yun, Tian Pei, Tianyu Sun, T.~Wang, Wangding Zeng, Wanjia Zhao, Wen Liu, Wenfeng Liang, Wenjun Gao, Wenqin Yu, Wentao Zhang, W.~L. Xiao, Wei An, Xiaodong Liu, Xiaohan Wang, Xiaokang Chen, Xiaotao Nie, Xin Cheng, Xin Liu, Xin Xie, Xingchao Liu, Xinyu Yang, Xinyuan Li, Xuecheng Su, Xuheng Lin, X.~Q. Li, Xiangyue Jin, Xiaojin Shen, Xiaosha Chen, Xiaowen Sun, Xiaoxiang Wang, Xinnan Song, Xinyi Zhou, Xianzu Wang, Xinxia Shan, Y.~K. Li, Y.~Q. Wang, Y.~X. Wei, Yang Zhang, Yanhong Xu, Yao Li, Yao Zhao, Yaofeng Sun, Yaohui Wang, Yi~Yu, Yichao Zhang, Yifan Shi, Yiliang Xiong, Ying He, Yishi Piao, Yisong Wang, Yixuan Tan, Yiyang Ma, Yiyuan Liu, Yongqiang Guo, Yuan Ou, Yuduan Wang, Yue Gong, Yuheng Zou, Yujia He, Yunfan Xiong, Yuxiang Luo, Yuxiang You, Yuxuan Liu, Yuyang Zhou, Y.~X. Zhu,
  Yanhong Xu, Yanping Huang, Yaohui Li, Yi~Zheng, Yuchen Zhu, Yunxian Ma, Ying Tang, Yukun Zha, Yuting Yan, Z.~Z. Ren, Zehui Ren, Zhangli Sha, Zhe Fu, Zhean Xu, Zhenda Xie, Zhengyan Zhang, Zhewen Hao, Zhicheng Ma, Zhigang Yan, Zhiyu Wu, Zihui Gu, Zijia Zhu, Zijun Liu, Zilin Li, Ziwei Xie, Ziyang Song, Zizheng Pan, Zhen Huang, Zhipeng Xu, Zhongyu Zhang, and Zhen Zhang.
\newblock Deepseek-r1: Incentivizing reasoning capability in llms via reinforcement learning, 2025.

\bibitem{schulman2017proximalpolicyoptimizationalgorithms}
John Schulman, Filip Wolski, Prafulla Dhariwal, Alec Radford, and Oleg Klimov.
\newblock Proximal policy optimization algorithms, 2017.

\bibitem{levine2020offlinereinforcementlearningtutorial}
Sergey Levine, Aviral Kumar, George Tucker, and Justin Fu.
\newblock Offline reinforcement learning: Tutorial, review, and perspectives on open problems, 2020.

\bibitem{fujimoto2019offpolicydeepreinforcementlearning}
Scott Fujimoto, David Meger, and Doina Precup.
\newblock Off-policy deep reinforcement learning without exploration, 2019.

\bibitem{andrychowicz2018hindsightexperiencereplay}
Marcin Andrychowicz, Filip Wolski, Alex Ray, Jonas Schneider, Rachel Fong, Peter Welinder, Bob McGrew, Josh Tobin, Pieter Abbeel, and Wojciech Zaremba.
\newblock Hindsight experience replay, 2018.

\bibitem{ecoffet2021goexplorenewapproachhardexploration}
Adrien Ecoffet, Joost Huizinga, Joel Lehman, Kenneth~O. Stanley, and Jeff Clune.
\newblock Go-explore: a new approach for hard-exploration problems, 2021.

\bibitem{wei2022finetunedlanguagemodelszeroshot}
Jason Wei, Maarten Bosma, Vincent~Y. Zhao, Kelvin Guu, Adams~Wei Yu, Brian Lester, Nan Du, Andrew~M. Dai, and Quoc~V. Le.
\newblock Finetuned language models are zero-shot learners, 2022.

\bibitem{sanh2022multitaskpromptedtrainingenables}
Victor Sanh, Albert Webson, Colin Raffel, Stephen~H. Bach, Lintang Sutawika, Zaid Alyafeai, Antoine Chaffin, Arnaud Stiegler, Teven~Le Scao, Arun Raja, Manan Dey, M~Saiful Bari, Canwen Xu, Urmish Thakker, Shanya~Sharma Sharma, Eliza Szczechla, Taewoon Kim, Gunjan Chhablani, Nihal Nayak, Debajyoti Datta, Jonathan Chang, Mike Tian-Jian Jiang, Han Wang, Matteo Manica, Sheng Shen, Zheng~Xin Yong, Harshit Pandey, Rachel Bawden, Thomas Wang, Trishala Neeraj, Jos Rozen, Abheesht Sharma, Andrea Santilli, Thibault Fevry, Jason~Alan Fries, Ryan Teehan, Tali Bers, Stella Biderman, Leo Gao, Thomas Wolf, and Alexander~M. Rush.
\newblock Multitask prompted training enables zero-shot task generalization, 2022.

\bibitem{chu2025sftmemorizesrlgeneralizes}
Tianzhe Chu, Yuexiang Zhai, Jihan Yang, Shengbang Tong, Saining Xie, Dale Schuurmans, Quoc~V. Le, Sergey Levine, and Yi~Ma.
\newblock Sft memorizes, rl generalizes: A comparative study of foundation model post-training, 2025.

\bibitem{ziegler2020finetuninglanguagemodelshuman}
Daniel~M. Ziegler, Nisan Stiennon, Jeffrey Wu, Tom~B. Brown, Alec Radford, Dario Amodei, Paul Christiano, and Geoffrey Irving.
\newblock Fine-tuning language models from human preferences, 2020.

\bibitem{luo2025empiricalstudycatastrophicforgetting}
Yun Luo, Zhen Yang, Fandong Meng, Yafu Li, Jie Zhou, and Yue Zhang.
\newblock An empirical study of catastrophic forgetting in large language models during continual fine-tuning, 2025.

\bibitem{kotha2024understandingcatastrophicforgettinglanguage}
Suhas Kotha, Jacob~Mitchell Springer, and Aditi Raghunathan.
\newblock Understanding catastrophic forgetting in language models via implicit inference, 2024.

\bibitem{fernando2025mitigatingforgettingllmsupervised}
Heshan Fernando, Han Shen, Parikshit Ram, Yi~Zhou, Horst Samulowitz, Nathalie Baracaldo, and Tianyi Chen.
\newblock Mitigating forgetting in llm supervised fine-tuning and preference learning, 2025.

\bibitem{xiong2025minimalistapproachllmreasoning}
Wei Xiong, Jiarui Yao, Yuhui Xu, Bo~Pang, Lei Wang, Doyen Sahoo, Junnan Li, Nan Jiang, Tong Zhang, Caiming Xiong, and Hanze Dong.
\newblock A minimalist approach to llm reasoning: from rejection sampling to reinforce, 2025.

\bibitem{mroueh2025reinforcementlearningverifiablerewards}
Youssef Mroueh.
\newblock Reinforcement learning with verifiable rewards: Grpo's effective loss, dynamics, and success amplification, 2025.

\bibitem{dong2024rlhfworkflowrewardmodeling}
Hanze Dong, Wei Xiong, Bo~Pang, Haoxiang Wang, Han Zhao, Yingbo Zhou, Nan Jiang, Doyen Sahoo, Caiming Xiong, and Tong Zhang.
\newblock Rlhf workflow: From reward modeling to online rlhf, 2024.

\bibitem{crowder2024hindsightexperiencereplayaccelerates}
Douglas~C. Crowder, Darrien~M. McKenzie, Matthew~L. Trappett, and Frances~S. Chance.
\newblock Hindsight experience replay accelerates proximal policy optimization, 2024.

\bibitem{wen2025lightr1curriculumsftdpo}
Liang Wen, Yunke Cai, Fenrui Xiao, Xin He, Qi~An, Zhenyu Duan, Yimin Du, Junchen Liu, Lifu Tang, Xiaowei Lv, Haosheng Zou, Yongchao Deng, Shousheng Jia, and Xiangzheng Zhang.
\newblock Light-r1: Curriculum sft, dpo and rl for long cot from scratch and beyond, 2025.

\bibitem{ouyang2022traininglanguagemodelsfollow}
Long Ouyang, Jeff Wu, Xu~Jiang, Diogo Almeida, Carroll~L. Wainwright, Pamela Mishkin, Chong Zhang, Sandhini Agarwal, Katarina Slama, Alex Ray, John Schulman, Jacob Hilton, Fraser Kelton, Luke Miller, Maddie Simens, Amanda Askell, Peter Welinder, Paul Christiano, Jan Leike, and Ryan Lowe.
\newblock Training language models to follow instructions with human feedback, 2022.

\bibitem{yang2025qwen3technicalreport}
An~Yang, Anfeng Li, Baosong Yang, Beichen Zhang, Binyuan Hui, Bo~Zheng, Bowen Yu, Chang Gao, Chengen Huang, Chenxu Lv, Chujie Zheng, Dayiheng Liu, Fan Zhou, Fei Huang, Feng Hu, Hao Ge, Haoran Wei, Huan Lin, Jialong Tang, Jian Yang, Jianhong Tu, Jianwei Zhang, Jianxin Yang, Jiaxi Yang, Jing Zhou, Jingren Zhou, Junyang Lin, Kai Dang, Keqin Bao, Kexin Yang, Le~Yu, Lianghao Deng, Mei Li, Mingfeng Xue, Mingze Li, Pei Zhang, Peng Wang, Qin Zhu, Rui Men, Ruize Gao, Shixuan Liu, Shuang Luo, Tianhao Li, Tianyi Tang, Wenbiao Yin, Xingzhang Ren, Xinyu Wang, Xinyu Zhang, Xuancheng Ren, Yang Fan, Yang Su, Yichang Zhang, Yinger Zhang, Yu~Wan, Yuqiong Liu, Zekun Wang, Zeyu Cui, Zhenru Zhang, Zhipeng Zhou, and Zihan Qiu.
\newblock Qwen3 technical report, 2025.

\bibitem{cobbe2021trainingverifierssolvemath}
Karl Cobbe, Vineet Kosaraju, Mohammad Bavarian, Mark Chen, Heewoo Jun, Lukasz Kaiser, Matthias Plappert, Jerry Tworek, Jacob Hilton, Reiichiro Nakano, Christopher Hesse, and John Schulman.
\newblock Training verifiers to solve math word problems, 2021.

\bibitem{yu2024metamathbootstrapmathematicalquestions}
Longhui Yu, Weisen Jiang, Han Shi, Jincheng Yu, Zhengying Liu, Yu~Zhang, James~T. Kwok, Zhenguo Li, Adrian Weller, and Weiyang Liu.
\newblock Metamath: Bootstrap your own mathematical questions for large language models, 2024.

\bibitem{wang2023planandsolvepromptingimprovingzeroshot}
Lei Wang, Wanyu Xu, Yihuai Lan, Zhiqiang Hu, Yunshi Lan, Roy Ka-Wei Lee, and Ee-Peng Lim.
\newblock Plan-and-solve prompting: Improving zero-shot chain-of-thought reasoning by large language models, 2023.

\bibitem{ye2025limoreasoning}
Yixin Ye, Zhen Huang, Yang Xiao, Ethan Chern, Shijie Xia, and Pengfei Liu.
\newblock Limo: Less is more for reasoning, 2025.

\bibitem{cheng2022hitabhierarchicaltabledataset}
Zhoujun Cheng, Haoyu Dong, Zhiruo Wang, Ran Jia, Jiaqi Guo, Yan Gao, Shi Han, Jian-Guang Lou, and Dongmei Zhang.
\newblock Hitab: A hierarchical table dataset for question answering and natural language generation, 2022.

\bibitem{deepscaler2024preview}
The~Agentica Team.
\newblock Deepscaler: Surpassing o1-preview with a 1.5b model by scaling rl, 2024.
\newblock Technical Report.

\end{thebibliography}
\bibliographystyle{unsrt}


\clearpage
\DoToC
\clearpage

\appendix

\section{Limitations and Future Work}
\label{sec:limitation}

While SuperRL demonstrates strong empirical performance across a variety of reasoning benchmarks, several limitations remain. First, the current fallback mechanism is reactive—it only resorts to supervised fine-tuning (SFT) when all sampled trajectories receive zero reward. While this avoids overwriting useful gradient signals from positive rollouts, it does not proactively leverage the offline data to shape early-stage exploration or mitigate suboptimal trajectory distributions. Incorporating more proactive or uncertainty-aware fallback strategies—such as reward prediction confidence, entropy thresholds, or offline pre-filtering—could further enhance learning stability.

Second, although SuperRL dynamically interleaves RL and SFT at the instance level, the two optimization objectives remain loosely coupled, alternating between distinct update rules. This decoupling may result in suboptimal credit assignment and hinder smoother gradient integration. A promising direction is to develop fully unified objectives or joint loss formulations that softly interpolate between SFT and RL signals, possibly guided by reward magnitude or rollout quality.

Third, our framework currently assumes access to high-quality expert trajectories for fallback SFT. However, such data may be unavailable or limited in some domains. Future work may explore bootstrapping fallback supervision from model self-refinements, synthetic generation, or retrieval-augmented reasoning.

Lastly, SuperRL is evaluated on static reasoning tasks with fixed inputs. Extending the framework to interactive or agentic environments—where reasoning is coupled with tool use, memory, or long-horizon planning—would require addressing new challenges in credit assignment, non-stationarity, and temporal abstraction. Moreover, integrating SuperRL with test-time search or planning methods (e.g., MCTS, beam-guided rollouts) remains an exciting avenue to push the limits of adaptive reasoning under sparse rewards.

Overall, while SuperRL provides a practical and effective solution to unify supervised and reinforcement-based reasoning, we believe it opens a broader design space for adaptive training paradigms that blend exploration, supervision, and search.

\section{Broader Impacts and Safeguards}
\label{sec:impacts}

As language models become increasingly capable of performing complex reasoning tasks, they are poised to influence high-stakes domains such as education, scientific discovery, legal analysis, and decision support. SuperRL, by improving the reasoning robustness of LLMs, could further accelerate this trend. However, such capabilities also introduce new societal risks, especially if models generate persuasive yet flawed reasoning or learn undesirable behaviors from biased feedback.

A key benefit of SuperRL lies in its fallback to high-quality supervised trajectories when reward signals are absent. This acts as a built-in safeguard, anchoring learning to human-verified reasoning paths and reducing the likelihood of reinforcement-induced divergence. However, reliance on offline data also inherits its limitations—such as annotation bias, narrow coverage, or stylistic homogenization—which could be amplified by repeated reuse. Careful curation, diversity analysis, and auditing of demonstration datasets are therefore crucial to mitigate these effects.

Moreover, since SuperRL adaptively integrates reinforcement learning, it inherits known risks from RL-based training: reward hacking, undesired generalization, and non-transparent credit assignment. Although our instance-level switching reduces over-optimization on spurious reward signals, stronger safeguards—such as adversarial evaluation, uncertainty estimation, or reward model interpretability—are needed to ensure alignment in more open-ended settings.

From a deployment perspective, reasoning models trained with SuperRL should be clearly scoped and audited before use in sensitive applications. Developers should monitor not only final accuracy but also behavioral changes introduced by the RL component, especially in failure cases. Human-in-the-loop evaluation, scenario red-teaming, and counterfactual probing can provide additional oversight.

Lastly, while SuperRL improves reasoning competence, it does not address deeper epistemic limitations of LLMs, such as lack of verifiability or causal grounding. Future safeguards may benefit from integrating structured verification, external tool use, or modular reasoning pipelines to augment the transparency and controllability of these systems.

\section{Detailed Settings of Experiments}
\label{ap:detail_setting}
\subsection{Dataset Configurations}

To rigorously evaluate \method, we utilize nine diverse datasets encompassing various reasoning challenges, from arithmetic problem-solving to competition-level mathematics and multi-agent reasoning. Below, we provide detailed descriptions of each dataset:

\paragraph{GSM8K} is a benchmark dataset comprising 8,500 high-quality grade school math word problems, crafted by human problem writers. The dataset is divided into 7,500 training and 1,000 test examples. Each problem typically requires 2 to 8 steps to solve, involving basic arithmetic operations such as addition, subtraction, multiplication, and division. GSM8K serves as a standard for evaluating multi-step mathematical reasoning in language models.

\paragraph{MetaMathQA} is a large-scale dataset containing 395,000 mathematical question-answer pairs. The dataset is generated by augmenting existing problems from GSM8K and MATH datasets, ensuring diversity and complexity in mathematical reasoning tasks. MetaMathQA aims to enhance the forward and backward reasoning capabilities of models across various mathematical domains, including algebra, geometry, and calculus.

\paragraph{PRM12K} is a subset of the PRM800K dataset, focusing on mathematical problems. It consists of 12,000 samples, each accompanied by five different solution paths, encompassing both correct and incorrect answers. Unlike some filtered datasets, PRM12K retains all samples, providing additional columns to indicate the correctness of each solution. This structure makes it particularly suitable for preference learning and reward modeling tasks.

\paragraph{OpenR1-Math-220k} is a comprehensive dataset designed for mathematical reasoning, comprising 220,000 math problems. Each problem is associated with two to four reasoning traces generated by the DeepSeek R1 model. The traces have been verified using tools like Math Verify and Llama-3.3-70B-Instruct, ensuring at least one correct reasoning path per problem. This dataset challenges models to understand and replicate complex reasoning processes.

\paragraph{LIMO} (Less is More for Reasoning) is a benchmark that challenges the conventional belief that large datasets are necessary for effective reasoning. It contains only 817 meticulously curated training samples, yet models trained on LIMO demonstrate superior performance across multiple benchmarks. LIMO emphasizes the quality of training data over quantity, showcasing that complex reasoning abilities can be elicited with limited but well-structured examples.

\paragraph{HiTab} is a dataset developed for question answering and natural language generation over hierarchical tables. It includes 3,597 tables and 10,686 QA pairs, sourced from statistical reports and Wikipedia pages. The tables exhibit complex hierarchical structures, and the dataset provides fine-grained annotations for entity and quantity alignment. HiTab poses significant challenges in numerical reasoning due to its hierarchical indexing and implicit semantic relationships.

\paragraph{AIME 2024} consists of 30 competition-grade problems from the 2024 AIME I and II exams. The dataset is formatted for open-ended reasoning with final boxed answers and is widely used in zero-shot evaluations. It focuses on symbolic, algebraic, and geometric reasoning under limited supervision.

\paragraph{AIME 2025} contains 30 problems from AIME I and II 2025, split into Part I and Part II. Compared to 2024, this version includes visual figures (e.g., TikZ-rendered diagrams), making it a more challenging benchmark for symbolic and spatial reasoning tasks. It is commonly evaluated using AveragePass@1.

\paragraph{DeepScaleR-Preview} is a curated subset of a larger multi-agent reasoning benchmark. The dataset emphasizes hierarchical reasoning across multiple steps, with annotations for subgoal structures. It spans mathematical, logical, and planning domains, and is designed to test both compositionality and scalability of reasoning models.

\begin{table}[ht]
\centering
\caption{Benchmarks used in this study. “--” indicates the split is not officially provided.}
\resizebox{\textwidth}{!}{%
\begin{tabular}{lcccccc}
\toprule
Dataset & \# Train & \# Test & Task Type & Domain & License & Source \\
\midrule
\textsc{GSM8K}~\cite{cobbe2021trainingverifierssolvemath} &
7{,}473 & 1{,}319 & Math word problems & Elementary math & MIT &
\href{https://huggingface.co/datasets/openai/gsm8k}{Link} \\

\textsc{MetaMathQA}~\cite{yu2024metamathbootstrapmathematicalquestions} &
395{,}000 & -- & Math QA & Mathematics & MIT &
\href{https://huggingface.co/datasets/meta-math/MetaMathQA}{Link} \\

\textsc{PRM12K}~\cite{wang2023planandsolvepromptingimprovingzeroshot} &
12{,}000 & -- & Programmatic reasoning & Mathematics & Apache 2.0 &
\href{https://huggingface.co/datasets/horseee/MixChain-Z-PRM12K}{Link} \\

\textsc{LIMO}~\cite{ye2025limoreasoning} &
817 & -- & Few-shot reasoning & Mathematics & MIT &
\href{https://github.com/GAIR-NLP/LIMO}{Link} \\

\textsc{OpenR1-Math-220k} &
220{,}000 & -- & Math reasoning & Mathematics & Apache 2.0 &
\href{https://huggingface.co/datasets/open-r1/OpenR1-Math-220k}{Link} \\

\textsc{HiTab}~\cite{cheng2022hitabhierarchicaltabledataset} &
7{,}399 & 1{,}583 & Hierarchical table QA & Statistics & C-UDA 1.0 &
\href{https://github.com/microsoft/HiTab}{Link} \\

\textsc{AIME 2024} &
-- & 30 & Competition math & Olympiad & MIT &
\href{https://huggingface.co/datasets/HuggingFaceH4/aime_2024}{Link} \\

\textsc{AIME 2025} &
-- & 30 & Competition math & Olympiad & CC BY 4.0 &
\href{https://huggingface.co/datasets/yentinglin/aime_2025}{Link} \\

\textsc{DeepScaleR-Preview}~\cite{deepscaler2024preview} &
-- & -- & Hierarchical reasoning & Mixed & Apache 2.0 &
\href{https://huggingface.co/datasets/agentica-org/DeepScaleR-Preview-Dataset}{Link} \\
\bottomrule
\end{tabular}}
\label{tab:reasoning_datasets}
\end{table}

\subsection{Model Configurations}

To comprehensively evaluate \method, we select a diverse set of open-source language models varying in size, architecture family, and training objectives. These models span three primary families: \textbf{Qwen2.5}, \textbf{LLaMA 3.x}, and \textbf{DeepSeek-R1 Distilled}. Below, we provide detailed descriptions of each:

\paragraph{Qwen2.5 Series} is a family of autoregressive language models developed by Alibaba, instruction-tuned for a wide range of reasoning tasks. It offers multiple model sizes—0.5B, 1.5B, 3B, and 7B—allowing systematic exploration of scale effects. All models are licensed under Apache 2.0 and trained with a consistent architecture design and tokenizer, ensuring comparability across sizes.

\paragraph{DeepSeek-R1-Distill-1.5B} is a distilled version of the original DeepSeek-R1 model, designed to preserve the reasoning capabilities of larger models while reducing computational overhead. It is instruction-tuned on diverse reasoning traces and optimized for both efficiency and generalization. This 1.5B model plays a key role in evaluating compact reasoning models.

\paragraph{LLaMA 3.x Series} includes models from the LLaMA 3.1 and 3.2 releases by Meta, with community-instruct fine-tuning. We use the 1B and 3B models from LLaMA 3.2, and the 8B model from LLaMA 3.1. These models are known for their competitive instruction-following ability and are widely adopted in the community for downstream alignment tasks.

All model weights are publicly available under permissive licenses, enabling reproducible benchmarking. Table~\ref{tab:models} summarizes the key properties of each model used in our experiments.

\begin{table}[ht]
\centering
\caption{Model configurations evaluated in our experiments.}
\resizebox{\textwidth}{!}{%
\begin{tabular}{lcccc}
\toprule
Model & Parameters (B) & Family & License & Source \\
\midrule
Qwen2.5-0.5B & 0.5 & Qwen2.5 & Apache 2.0 &
\href{https://github.com/QwenLM/Qwen2.5}{Link} \\
Qwen2.5-1.5B & 1.5 & Qwen2.5 & Apache 2.0 &
\href{https://github.com/QwenLM/Qwen2.5}{Link} \\
Qwen2.5-3B & 3 & Qwen2.5 & Apache 2.0 &
\href{https://github.com/QwenLM/Qwen2.5}{Link} \\
Qwen2.5-7B & 7 & Qwen2.5 & Apache 2.0 &
\href{https://github.com/QwenLM/Qwen2.5}{Link} \\
R1-Distill-1.5B & 1.5 & DeepSeek-R1 & MIT &
\href{https://huggingface.co/deepseek-ai}{Link} \\
Llama3.2-1B-Instruct & 1 & Llama 3.2 & Llama 3 Community &
\href{https://huggingface.co/meta-llama/Llama-3.2-1B}{Link} \\
Llama3.2-3B-Instruct & 3 & Llama 3.2 & Llama 3 Community &
\href{https://huggingface.co/meta-llama/Llama-3.2-3B}{Link} \\
Llama3.1-8B-Instruct & 8 & Llama 3.1 & Llama 3 Community &
\href{https://huggingface.co/meta-llama/Llama-3.1-8B}{Link} \\
\bottomrule
\end{tabular}}
\label{tab:models}
\end{table}

\subsection{Construction and Utilization of Offline Reasoning Data}

The offline SFT data utilized in \method is derived from expert-annotated or high-quality model-generated reasoning traces. Many of the datasets we employ inherently contain rich, step-by-step problem-solving trajectories, serving as natural sources of offline supervision.

For instance, the GSM8K dataset provides detailed answer fields comprising multi-step reasoning chains and final solutions, making it directly suitable for use as SFT targets. Similarly, PRM12K offers both expert-annotated solutions and a substantial portion of high-quality, model-generated reasoning traces. We meticulously validate and include these generated traces in our SFT corpus when their final answers align with the correct labels.

LIMO, originally designed to study long-horizon mathematical reasoning, was explicitly structured to benefit from supervised learning. Its strong performance under SFT alone motivated our inclusion of LIMO in the hybrid training setting, leveraging its expert-annotated solutions as high-quality offline supervision signals.

In contrast, HiTab lacks annotated intermediate reasoning steps, which are crucial for supervised learning in our hybrid training framework. To address this, we generate synthetic reasoning trajectories using a large instruction-tuned model (e.g., Deepseek-R1), prompted to produce step-by-step justifications for each QA pair. We implement a stringent answer-filtering mechanism: only when the generated rationale's final answer matches the ground-truth label is it accepted into the SFT corpus. This ensures high precision and prevents low-quality traces from corrupting training.

\paragraph{Implementation in \textsc{VeRL}.}

Within the \textsc{VeRL} framework, whose license is Apache License 2.0, we adopt a structured approach to integrate offline trajectories into the training pipeline:

\begin{itemize}
\item \textbf{Data Annotation and Storage}: Each data sample is augmented with an extra info field, encapsulating metadata such as the original question, the extracted or generated reasoning trajectory, and the target answer. This design ensures that auxiliary information is preserved alongside the primary data, facilitating downstream processing.
\item \textbf{Custom Dataset Class}: We define a custom \texttt{Dataset} class that preprocesses the annotated data. This class is responsible for converting each entry into a standardized \texttt{DataProto} object. The \texttt{DataProto} includes tokenized inputs, loss masks for supervised targets, and auxiliary fields for logging and analysis. This modular design promotes flexibility and reusability across different training configurations.
\item \textbf{Integration with Actor Module}: The \texttt{Actor} module accesses the preprocessed \texttt{DataProto} objects during training. By leveraging the structured information within each \texttt{DataProto}, the \texttt{Actor} can efficiently retrieve high-quality supervision signals, ensuring that the model benefits from the rich reasoning trajectories during optimization.
\end{itemize}

This integration strategy ensures that the offline trajectories are seamlessly incorporated into the hybrid optimization process, providing consistent behavioral priors and enhancing the model's reasoning capabilities.

\subsection{Prompt and Output Format Design}
\label{sec:prompt_design}

To promote interpretable and verifiable reasoning behavior, we adopt distinct prompting and output formatting strategies tailored to the model type.

\paragraph{Prompt Design.}
For both vanilla and structured-output models, we append a concise instruction—``Let's think step by step and output the final answer in \texttt{boxed\{\}}.''—to the original problem description to encourage step-by-step reasoning and a clearly identifiable final answer.

For example:
\begin{quote}
\textbf{Prompt:} If you have 3 apples and you buy 2 more, how many do you have? Let's think step by step and output the final answer in boxed\{\}.
\end{quote}

For \textbf{vanilla models} without structured output conventions, the model is directly given the above prompt string. A typical output is:

\begin{quote}
\texttt{We start with 3 apples. Buying 2 more gives us 3 + 2 = 5. The final answer is boxed\{5\}.}
\end{quote}

In contrast, for \textbf{structured-output models} (e.g., Qwen2.5-1.5B, DeepSeek-R1-Distill-Qwen-1.5B) that support chat-style prompting, we apply \texttt{apply\_chat\_template} to transform the prompt into a ChatML-formatted conversation. For instance:

\begin{quote}
\texttt{<|im\_start|>system\\
You are a helpful assistant.<|im\_end|>\\
<|im\_start|>user\\
If you have 3 apples and you buy 2 more, how many do you have? Let's think step by step and output the final answer in boxed\{\}.<|im\_end|>\\
<|im\_start|>assistant\\
<think>}
\end{quote}

\paragraph{Expected Model Output and Postprocessing.}
The output format likewise depends on the model's interface and training. For structured-output models, the expected output includes special tags:

\begin{quote}
\texttt{We start with 3 apples. Buying 2 more gives us 3 + 2 = 5. Now let's output the final answer.</think>\\
<answer>The answer is boxed\{5\}</answer>}
\end{quote}

In this case, we apply tag-aware parsing during postprocessing:
\begin{itemize}
    \item Extract the reasoning trace enclosed in \texttt{<think>} tags.
    \item Extract the final answer from within the \texttt{<answer>} tag, specifically the content of \texttt{boxed\{\}}.
\end{itemize}

For vanilla models, which do not include tags, we rely on regex-based postprocessing:
\begin{itemize}
    \item Identify the reasoning portion as any content before the appearance of \texttt{boxed\{\}}.
    \item Parse the numerical answer from within \texttt{boxed\{\}}.
\end{itemize}

These model-aware prompt formatting and output parsing strategies ensure the consistent interpretation of model responses across evaluation and training pipelines. They also enable structured reward computation and answer matching for reinforcement learning optimization.

\subsection{Reward Design}
\label{sec:reward_design}

Given a model response $y$ to input $x$, we compute the reward as a binary signal:
\[
r(x, y) =
\begin{cases}
1 & \text{if } \text{extract}(y) = y^{\text{gt}} \\
0 & \text{otherwise}
\end{cases}
\]
where $\text{extract}(y)$ denotes the parsed answer obtained via model-specific extraction logic, and $y^{\text{gt}}$ is the ground-truth answer.

To accommodate surface-form variations—especially prevalent in datasets such as LIMO and PRM12K—we incorporate a \emph{canonicalization layer} in the reward computation. This module standardizes answer representations by:
\begin{itemize}
    \item Normalizing numeric formats (e.g., converting fractions to decimals);
    \item Performing symbolic equivalence checks (e.g., $2x + 4$ vs.\ $4 + 2x$);
    \item Unifying variable names, units, or other context-specific notations.
\end{itemize}
If the canonicalized prediction matches any canonicalized gold reference, we assign a reward of 1.

As a fallback mechanism, when no delimiters (e.g., \texttt{\textbackslash boxed\{\}}) are detected in the model output, we extract the last numerical span as a proxy for the final answer. All reward functions are implemented as deterministic, stateless modules to ensure reproducibility and compatibility with batched rollout evaluations in PPO and GRPO training pipelines.

\subsection{Metric Design}
\label{sec:metric_design}

We adopt \textbf{Exact Match (EM)} accuracy as the primary evaluation metric to assess model performance on reasoning tasks. A prediction is considered correct if the extracted answer exactly matches the ground-truth answer after normalization. This includes removing extraneous formatting, standardizing numerical representations, and optionally applying symbolic simplification when applicable. EM offers a strict yet interpretable signal of end-to-end correctness, effectively capturing whether the model arrives at the correct final solution.

Compared to token-level metrics such as BLEU or ROUGE—which quantify n-gram overlap—EM is more aligned with the discrete nature of most reasoning tasks. Token-based metrics often tolerate superficial similarity while overlooking semantically crucial deviations (e.g., predicting $7.0$ instead of $7.5$), thus failing to penalize incorrect answers that appear linguistically similar. In contrast, EM enforces a high bar for correctness by requiring exact alignment with the reference answer.

To accommodate datasets with multiple valid reasoning paths or equivalent solutions—such as PRM12K and OpenR1—we extend EM to a \emph{relaxed matching} scheme. Specifically, a prediction is marked as correct if it matches \emph{any} of the acceptable reference answers after canonicalization. This allows for flexibility in surface forms (e.g., equivalent algebraic expressions or unit conversions) while preserving the core requirement of semantic equivalence.

In all cases, the normalization and canonicalization procedures used during EM evaluation are kept deterministic and model-agnostic to ensure reproducibility and fairness. This design choice ensures that the metric remains robust across diverse model architectures and output formats.

\subsection{Environment Setup}
\label{appendix:environment}

All experiments are conducted within the \texttt{verl} framework, a scalable actor–critic reinforcement learning platform tailored for optimizing language models. This framework serves as the foundation for our experimental setup, allowing us to implement and iterate on various training strategies. Our primary experiments utilize GRPO as the core reinforcement learning algorithm. However, to validate the general applicability of our uncertainty-weighted hybrid training framework, we also conduct comparative trials using PPO. The configurations, scripts and source code for these experiments are all developed and modified within the \texttt{verl}. This includes the implementation of the GRPO and PPO algorithms, as well as the integration of the uncertainty-weighted hybrid training approach. The flexibility of the \texttt{verl} enables us to seamlessly update and refine our methods, ensuring that our experiments are both robust and adaptable.

Our experimental scripts follow a unified and modular configuration framework designed to ensure consistency across all datasets and model backbones. This framework supports flexible adaptation, with minor adjustments made to accommodate variations in model scale (e.g., parameter count) and available computational resources (e.g., GPU memory capacity). By default, both the actor and critic are initialized from the same pretrained checkpoint. This shared initialization helps maintain stability in the early stages of training and ensures that policy updates build upon a consistent starting point. Unless otherwise noted, we train the entire model with full-parameter updates, rather than using techniques like partial tuning or adapters. To manage memory consumption during training, we enable gradient checkpointing, which trades off additional computation for significantly reduced memory usage. This is particularly important when training large models with long sequences or large batch sizes. To constrain policy drift and encourage stable learning, we apply KL divergence regularization between the current policy and a fixed reference policy. We use a low-variance KL formulation with a fixed regularization coefficient of 0.001, following prior work showing its effectiveness in language model fine-tuning. 

We adopt a fixed learning rate of 1e-6 for all experiments, regardless of dataset or model size. This consistent setting simplifies hyperparameter tuning and facilitates fair comparisons across different tasks. The batch size is set to 32 by default, which provides a good balance between training stability and GPU memory efficiency. However, in data-scarce scenarios—such as the \textsc{LIMO} dataset, which contains relatively few high-quality training samples—we reduce the batch size to 8 to improve gradient quality and mitigate overfitting risks associated with small datasets. Each training run proceeds for 500 update steps, a schedule empirically chosen to ensure sufficient optimization while maintaining computational efficiency. To monitor progress and detect training instabilities early, we conduct model evaluation every 5 steps on a held-out validation set. At the end of training, we report the test-set performance at step 500 if the learning curve shows smooth convergence. In cases where the validation curve exhibits fluctuations or noise—typically due to reward sparsity or instability in policy gradients—we apply exponential moving average (EMA) smoothing to the score trajectory to obtain a more reliable final evaluation metric. This ensures that our reported performance reflects the overall trend rather than being biased by momentary spikes or drops.

For rollout generation, we employ the \texttt{vLLM} backend, which enables efficient and scalable batched decoding with optimized GPU memory usage. Each prompt is decoded to produce five candidate responses, allowing for diverse sampling during training. To ensure stable runtime behavior, we cap GPU memory utilization at 40\%. The decoding configuration adopts a stochastic sampling strategy with a temperature of 1.0, top-$k = -1$ (disabled), and top-$p = 1.0$, corresponding to unconstrained nucleus sampling. These settings encourage diverse yet coherent response generation from the model. We constrain the maximum sequence length—comprising both the input prompt and the generated output—to 2048 tokens. Prompts that exceed this limit are filtered out prior to generation, and any attempt to exceed the limit during decoding raises a truncation error. This strict enforcement helps maintain consistency and prevents the introduction of ambiguous or malformed training signals.

For dataset partitioning, we follow the original train/test splits provided by the benchmark for \textsc{GSM8K} and \textsc{HiTab}, ensuring compatibility with prior work and fair comparison. For all other datasets containing more than 20{,}000 examples, we randomly subsample a total of 20{,}000 instances to reduce computational cost while maintaining representative coverage. The selected subset is then split into training and test sets using an 80/20 ratio. This standardized partitioning protocol facilitates consistent evaluation across diverse datasets with varying sizes and distributions.

All experiments are conducted on machines equipped with NVIDIA H100 GPUs. For most settings involving smaller models (e.g., 1-3B parameters), we utilize a 2-GPU configuration, which provides sufficient compute capacity for full-parameter training. In contrast, larger models (e.g., 7B and above) are trained in a distributed fashion using multiple nodes and GPUs. We construct multi-node clusters using \texttt{Ray}, a flexible framework for large-scale distributed computing. Once initialized, training is orchestrated through \texttt{verl}'s distributed utilities, which support scalable actor–critic reinforcement learning with efficient inter-GPU communication and synchronization. The micro-batch size is fixed at 2 per GPU across all experiments, regardless of the number of nodes or model size. This setting balances memory usage and gradient estimation stability, especially under reinforcement learning with sparse rewards. To ensure a fair comparison across different training paradigms, we adopt a unified optimizer configuration—including learning rate, weight decay, and scheduler—for all methods: supervised-only (SFT), reinforcement learning-only (RL), sequential SFT+RL, and our proposed hybrid RL+SFT. This design isolates the effect of training strategies from confounding optimization differences.

For the \textsc{SFT} and \textsc{SFT+RL} baselines, we begin by fine-tuning the base model using the same training dataset, learning rate, and context length as in the RL-based training setups. The supervised fine-tuning is conducted for a total of 25 epochs, a duration chosen to ensure sufficient convergence without overfitting. Throughout training, we evaluate each epoch's checkpoint on the held-out test set using greedy decoding, i.e., with temperature set to zero and no sampling. The checkpoint achieving the highest test-set performance is selected as the final \textsc{SFT} baseline. For the \textsc{SFT+RL} baseline, reinforcement learning is initialized from this best-performing \textsc{SFT} checkpoint. We then continue training the model under the same RL framework used in our hybrid method. The final performance of the \textsc{SFT+RL} model is reported based on the test-set score at the point where the learning curve reaches stable convergence. If the learning dynamics are noisy, we apply exponential moving average (EMA) smoothing to determine the final score in a robust manner.

\section{Lessons Learned from Failed Attempts and Design Iterations}
\label{sec:lessions}
Throughout the development of \method, we explored a range of alternative strategies aimed at improving data efficiency and stability in RL for language models. A key motivating principle behind our early design iterations was to fully leverage existing supervision data within the RL training loop, regardless of whether it was collected off-policy or generated dynamically. This led us to investigate several hybrid approaches that sought to maximize the utility of available data across both supervised and reinforcement learning signals. Despite their conceptual appeal, many of these attempts failed to yield meaningful improvements, and in some cases, actively degraded performance. Below, we summarize the most instructive failures.

\subsection{Offline Data as GRPO-Compatible Rollouts: A Negative Result}

To investigate the potential of incorporating static expert supervision within GRPO training, we augment each sampled prompt by treating expert-annotated responses as auxiliary rollouts. Concretely, for every on-policy prompt sampled during training, the corresponding offline trajectory is injected into the GRPO buffer with its reward computed via a task-specific metric (e.g., exact match). These expert rollouts coexist with policy-generated samples, contributing to the surrogate loss and gradient updates. This design aims to regularize policy optimization and accelerate convergence by leveraging high-quality behavioral signals.

\textbf{Empirical Collapse.} Despite these motivations, both rollout-injection variants led to catastrophic learning failure: validation accuracy declined consistently throughout training, indicating that static expert rollouts destabilized optimization and accelerated mode collapse.

\subsubsection{Variant I: Direct Injection of Expert Rollouts}

This variant treats offline expert trajectories as if they were on-policy samples from the current policy $\pi_\theta$, directly incorporating them into the GRPO buffer for joint optimization. At each training step, we sample a batch of expert trajectories $(x_e, y_e) \sim \mathcal{D}_{\text{expert}}$, assign rewards $r(x_e, y_e)$ based on task-specific metrics, and compute the mixed surrogate loss:
\begin{equation}
\mathcal{L}_{\text{mixed}} = \mathbb{E}_{(x, y) \sim \pi_\theta} \left[ w(x, y) \cdot \log \pi_\theta(y|x) \right] + \lambda \cdot \mathbb{E}_{(x_e, y_e) \sim \mathcal{D}_{\text{expert}}} \left[ r(x_e, y_e) \cdot \log \pi_\theta(y_e|x_e) \right],
\label{eq:direct_as_rollout}
\end{equation}
where the first term corresponds to preference-weighted GRPO rollouts, and the second term injects expert supervision weighted by scalar $\lambda$.

\textbf{Observed Failure.} This strategy consistently degraded performance across all benchmarks (e.g., LIMO, HiTab), slowing convergence and increasing KL divergence relative to both pure GRPO and reward-aware rollout-to-SFT methods. We attribute this to \emph{distributional mismatch}: expert trajectories $(x_e, y_e)$ are not sampled from $\pi_\theta$, violating GRPO’s on-policy assumption and introducing high-variance gradients. As a result, the policy oscillates between chasing unreachable expert targets and reinforcing suboptimal behaviors. This is reflected in elevated KL divergence:
\begin{equation}
\text{KL}(\pi_\theta \| \pi_{\text{ref}}) = \mathbb{E}_{x} \left[ \sum_{y} \pi_\theta(y|x) \log \frac{\pi_\theta(y|x)}{\pi_{\text{ref}}(y|x)} \right],
\end{equation}
signaling instability and misalignment with the initial reference distribution. Furthermore, static expert rollouts fail to adapt to the evolving policy’s exploration frontier, providing little guidance in unexplored regions of the solution space.

\subsubsection{Variant II: Self-Rewritten Expert Rollouts}

To address the distributional mismatch, we introduce a \emph{self-rewrite} mechanism wherein expert responses are reformulated by the current policy to ensure alignment with its own distribution. Given $(x_e, y_e) \sim \mathcal{D}_{\text{expert}}$, we prompt $\pi_\theta$ to generate a rewritten response $\hat{y}_e \sim \pi_\theta(\cdot\,|\,x_e, y_e)$ that is semantically equivalent to $y_e$. The rewritten trajectory $(x_e, \hat{y}_e)$ is assigned the same reward as the original expert response, yielding the loss:
\begin{equation}
\mathcal{L}_{\text{rewrite}} = \mathbb{E}_{(x, y) \sim \pi_\theta} \left[ w(x, y) \cdot \log \pi_\theta(y|x) \right] + \lambda \cdot \mathbb{E}_{(x_e, \hat{y}_e) \sim \pi_\theta(\cdot\,|\,x_e, y_e)} \left[ r(x_e, \hat{y}_e) \cdot \log \pi_\theta(\hat{y}_e | x_e) \right].
\label{eq:rewrite_loss}
\end{equation}

\textbf{Observed Failure.} While distributional alignment is improved, this method failed to yield performance gains. Rewritten responses often diverged semantically from expert outputs, especially in multi-step reasoning tasks, introducing \emph{semantic drift}. Assigning expert-level rewards $r(x_e, \hat{y}_e) = r(x_e, y_e)$ to imperfect rewrites led to reward misalignment:
\begin{equation}
\mathbb{E}[r(x_e, \hat{y}_e)] < r(x_e, y_e), \quad \text{but used reward } \approx r(x_e, y_e).
\end{equation}
This misalignment inflated value predictions and reinforced spurious reasoning paths, resulting in cumulative confirmation bias and unstable convergence.

\subsection{Conclusion: Offline Rollouts Undermine GRPO Stability}

Our findings reveal that treating offline trajectories as GRPO-compatible rollouts—either directly or via self-rewriting—fails to improve and often degrades performance. These strategies violate GRPO’s on-policy assumptions and introduce high-variance or semantically misaligned gradients. We conclude that more principled methods are needed to bridge the gap between static supervision and dynamic exploration without compromising the stability of reinforcement learning updates.

\subsection{Offline Data as Few-shots}
To more effectively leverage the information contained in high-quality offline data, we explored augmenting the original prompt by prepending additional exemplars drawn from the same supervised dataset. These few-shot exemplars, comprising other question-solution pairs, are intended to serve as implicit guidance. By exposing the model to a wider range of reasoning patterns, we aim to enhance its generalization capabilities.

We explored three selection strategies for constructing these few-shot demonstrations: \textbf{Random Selection:} Randomly sample one or more exemplars from the SFT dataset without conditioning on the current prompt. \textbf{Prompt Similarity:} Retrieve exemplars whose questions are semantically similar to the target prompt, based on embedding-based similarity. \textbf{Solution Similarity:} Select exemplars whose solutions are structurally or semantically similar to the expert solution of the target prompt.

For each strategy, we varied the number of exemplars included in the prompt (one vs. three) to test whether increasing few-shot context leads to measurable improvements.

\begin{table}[t!]
\centering
\caption{Few-shot Prompting Performance with \textbf{Random Selection} (Qwen2.5-1.5B, GRPO)}
\label{tab:fewshot_random_summary}
\resizebox{\textwidth}{!}{
\begin{tabular}{lcccccl}
\toprule
\textbf{Dataset} & \textbf{Vanilla} & \textbf{1-shot (Random)} & \textbf{3-shot (Random)} & \textbf{Best (Random)} & \textbf{Gain} & \textbf{Observation} \\
\midrule
GSM8K    & 0.755  & 0.747  & 0.745  & \textbf{Vanilla} & \textcolor{red}{-0.010} & Few-shots degrade performance \\
MetaMath & 0.814  & 0.793  & 0.8255 & \textbf{3-shot} & \textcolor{green!60!black}{+0.011} & Faster convergence; slight gain \\
PRM12K   & 0.505  & 0.480  & 0.5057 & \textbf{3-shot} & \textcolor{green!60!black}{+0.007} & Slight gain \\
LIMO     & 0.018  & 0.018 $\rightarrow$ 0 & 0.018 $\rightarrow$ 0 & \textbf{None} & \textcolor{red}{Collapse} & All models collapse to 0 \\
\bottomrule
\end{tabular}
}
\end{table}

\textbf{Empirical Observations.} Despite the intuitive appeal of enriching the prompt with related examples, none of the strategies consistently outperformed the base model across benchmarks. The effectiveness of few-shot augmentation was highly variable: while minor gains were observed on specific reasoning-centric datasets such as MetaMath and PRM12K, performance degraded or remained stagnant on others, including GSM8K and LIMO. Notably, LIMO exhibited complete collapse under few-shot augmentation, echoing similar failure patterns seen in other unstable training regimes.

We performed extensive evaluations using the \textbf{Random Select} strategy across four datasets, while the \textbf{Prompt Similarity} and \textbf{Solution Similarity} strategies were validated on a subset due to computational constraints. Nevertheless, all methods displayed consistent patterns: no significant improvement was achieved over the vanilla setting, and the inclusion of static exemplars occasionally introduced additional variance or learning instability. Crucially, none of the approaches achieved the type of meaningful gains observed under our proposed hybrid training strategy, which dynamically balances supervised and policy gradient objectives.

In summary, our findings highlight the limitations of offline few-shot augmentation as a plug-and-play solution for improving GRPO training. The static nature of these exemplars, combined with potential mismatches in reasoning complexity or style, undermines their utility as general-purpose inductive scaffolds. While few-shot prompting may offer minor benefits in specific cases, it fails to deliver consistent performance gains and does not constitute a reliable substitute for hybrid or dynamically supervised training methods.

\section{Training Graph Analysis}
\begin{figure}[ht]
    \centering
    \includegraphics[width=0.95\linewidth]{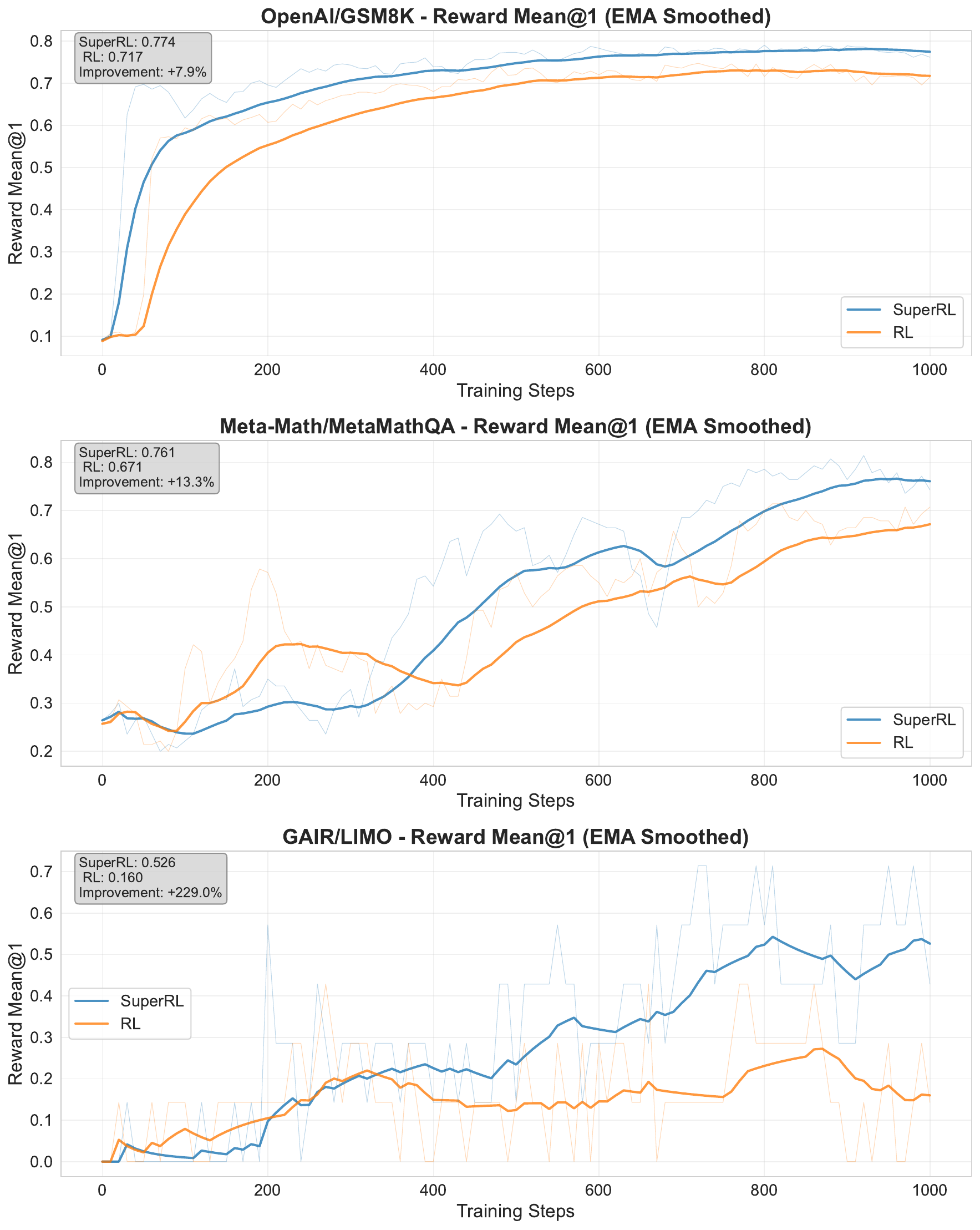}
    \caption{Training curve on the GSM8K dataset with \textbf{dense reward}. This graph shows how the performance evolves over time under sparse reward environment.}
    \label{fig:training_gsm8k}
\end{figure}

\begin{figure}[ht]
    \centering
    \includegraphics[width=0.95\linewidth]{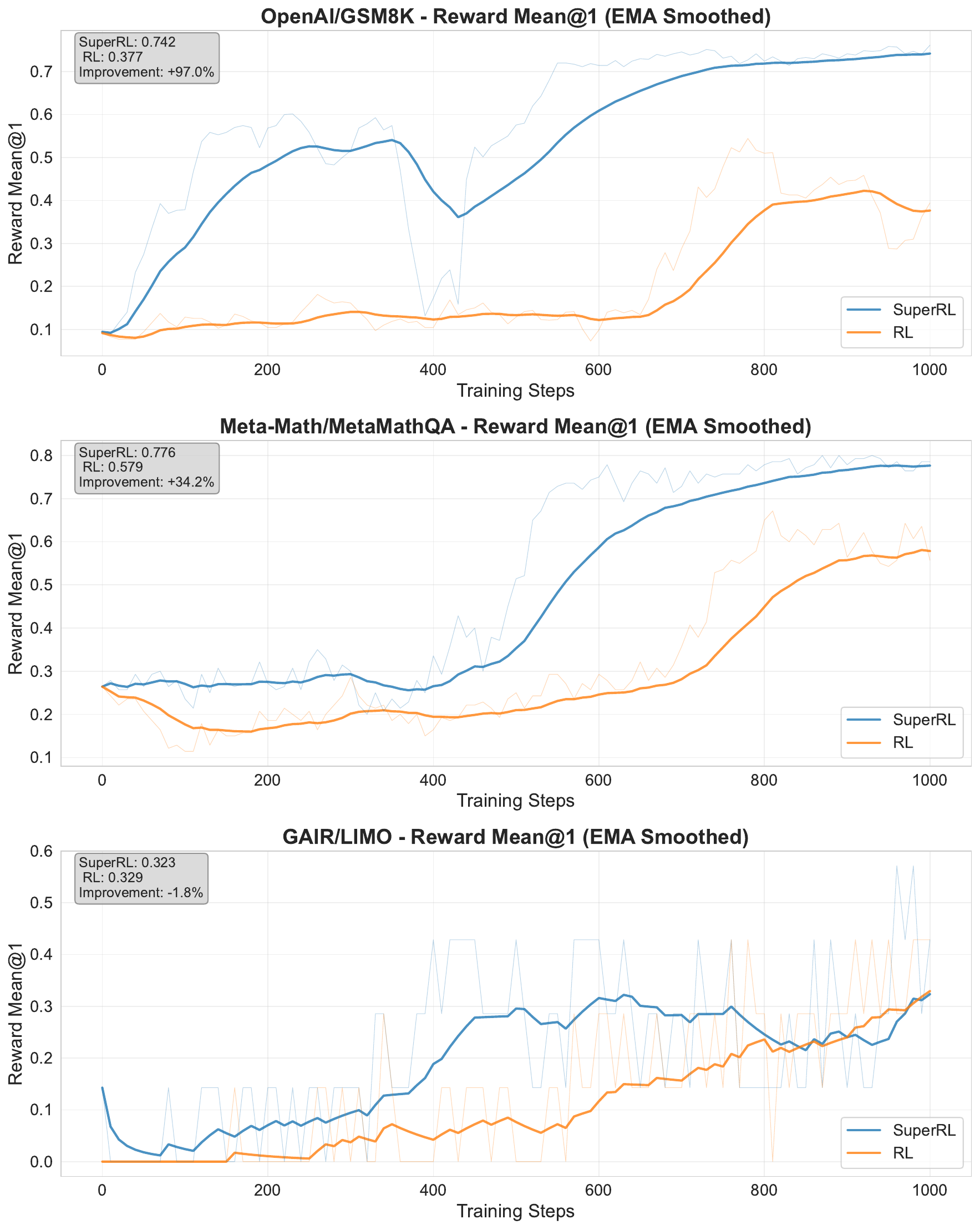}
    \caption{Training curve on the OpenR1 dataset with \textbf{sparse reward}. This graph illustrates the challenges and learning dynamics under sparse supervision.}
    \label{fig:training_openr1}
\end{figure}

The training dynamics depicted in \ref{fig:training_gsm8k} and \ref{fig:training_openr1} offer detailed insights into how SuperRL and RL perform across three datasets under distinct reward frameworks, with all metrics tracked as EMA-smoothed Reward Mean@1 over training steps.
In GSM8KTrainingGraph.pdf, which focuses on dense reward scenarios, SuperRL demonstrates a clear and consistent advantage over RL across all datasets. On OpenAI/GSM8K, SuperRL reaches a final Reward Mean@1 of 0.774, marking a 7.9\% improvement compared to RL’s 0.717. The curves show a steady divergence as training progresses, indicating that SuperRL effectively harnesses the dense reward signals to refine its performance over time. For Meta-Math/MetaMathQA, the gap is more pronounced: SuperRL achieves 0.761, a 13.3\% gain over RL’s 0.671, suggesting its strong adaptability to mathematical reasoning tasks where dense feedback provides critical step-by-step guidance. Most notably, on GAIR/LIMO, SuperRL delivers a remarkable 229.0\% improvement, with a final value of 0.526 versus RL’s 0.160, underscoring its ability to thrive in complex reasoning environments with dense reward structures.

Turning to \ref{fig:training_openr1}, which examines sparse reward settings, SuperRL’s performance shows nuanced strengths and a single exception. On OpenAI/GSM8K, it achieves a striking 97.0\% improvement, with a final Reward Mean@1 of 0.742 compared to RL’s 0.377, highlighting its proficiency in extracting meaningful signals from limited feedback in this dataset. For Meta-Math/MetaMathQA, SuperRL maintains a solid edge, reaching 0.776 (34.2\% higher than RL’s 0.579), with sustained separation in the curves indicating robust learning under sparse supervision. However, on GAIR/LIMO, SuperRL slightly underperforms, with a final value of 0.323 (-1.8\% compared to RL’s 0.329), suggesting that the dataset’s unique characteristics, when combined with sparse rewards, create a scenario where RL’s optimization approach aligns more effectively with the limited feedback.

Together, these graphs from \ref{fig:training_gsm8k} and \ref{fig:training_openr1} illustrate that SuperRL generally excels in both dense and sparse reward environments, with its performance gains being particularly significant in dense settings and most sparse scenarios, while also revealing a specific case where RL holds a marginal advantage.

\section{Soft Fusion over Hard Switching: A Unified Perspective on SuperRL versus SFT+RL}
\label{appendix:soft-vs-hard}

A core challenge in aligning large language models (LLMs) with complex reasoning behaviors lies in how to effectively combine supervised fine-tuning (SFT) and reinforcement learning (RL). While SFT offers high-quality, human-aligned demonstrations, RL allows the model to adapt toward task-specific rewards and discover novel, high-utility behaviors. Yet integrating these two signals remains nontrivial: naïve combinations often result in instability, forgetting, or inefficiency.

The traditional SFT+RL framework adopts a \textit{hard switching} strategy. It first optimizes the model using offline expert data via SFT, and then transitions to on-policy RL to maximize downstream reward signals. While simple and modular, this two-stage approach suffers from critical limitations. First, the abrupt shift in objectives often causes \textit{catastrophic forgetting} of previously learned behavior, especially under sparse or noisy rewards. Second, once the model is anchored to demonstration-like behaviors, it may resist exploring off-distribution actions that yield higher long-term reward. Third, RL without continual alignment can overfit to narrow reward functions, producing undesired or brittle behavior.

In contrast, the latest version of \textbf{SuperRL} implements a \textit{lightweight but principled soft fusion} strategy: for each training instance, the model performs multiple rollouts and examines their rewards. If any rollout receives a non-zero reward, RL updates are applied; otherwise, the model falls back to supervised updates using expert demonstrations. This dynamic instance-level switching seamlessly blends exploration with imitation: rather than predefining a training schedule or manually tuning loss weights, SuperRL adaptively selects the most informative signal—RL or SFT—for each input based on reward feedback.

This formulation brings several practical and conceptual advantages:
\begin{itemize}
    \item It avoids sharp transitions between learning paradigms and retains the \textit{stability of SFT} throughout training.
    \item It ensures that \textit{online exploration is grounded by offline knowledge} in low-reward regions, improving learning in sparse-reward or hard-exploration regimes.
    \item It introduces no additional architectural complexity—only a reward-gated control flow—making it both simple to implement and scalable across tasks and model sizes.
    \item It encourages \textit{generalization beyond demonstrations} by using SFT only when needed, preventing the model from overfitting to static traces.
\end{itemize}

Compared to previous “soft fusion” approaches that require manually tuned loss interpolation or gradient-level blending, SuperRL leverages reward feedback as a \textit{data-driven switch}, reducing the risk of signal interference while preserving training efficiency. This design effectively bridges the strengths of both paradigms: \textit{SFT provides a fallback anchor when RL fails, and RL enables adaptive generalization when reward signals are informative}.

Nonetheless, SuperRL’s strategy is not without limitations. First, it relies on \textit{reward observability}: when rewards are extremely noisy or delayed, the switch decision may become unstable. Second, unlike continuous fusion, the fallback mechanism may not fully capture finer-grained trade-offs between imitation and exploration within the same instance. Third, in domains where reward functions are dense and fully aligned with demonstrations (e.g., synthetic games), pure RL may suffice and converge faster without SFT fallback.

Even so, for open-ended reasoning, long-horizon inference, and weak supervision settings—where reward signals are partial and data distributions shift—SuperRL offers a \textit{robust, generalizable, and efficient alternative to hard-switch SFT+RL pipelines}. It avoids the brittleness of static methods while maintaining a clean training loop and strong empirical performance.

In summary, SuperRL reimagines soft fusion not as continuous loss interpolation but as \textit{instance-level reward-aware fallback}, enabling practical and scalable integration of offline supervision with online exploration. This simple yet effective strategy offers a compelling alternative to traditional hard-switching pipelines, particularly for real-world LLM training under sparse or uncertain feedback.

\end{document}